# The Geometry of LLM-as-Judge: Why Inter-LLM Consensus Is Not Human Alignment

**Sourabrata Mukherjee Hamna Hamna Kalika Bali Sunayana Sitaram**
Microsoft Research
t-somukherje@microsoft.com

## Abstract

LLM judges now score most open-ended NLP output, and their mutual agreement is routinely read as evidence that the scores can be trusted. That reading is unsafe: judges may agree because they capture quality, or because they share the same blind spots, and agreement statistics alone cannot tell these apart. We develop a geometric test that can. Treating each judge's scores as a vector, we measure spread, effective rank, the angle to human scores, and the judge–judge, judge–human, and human–human agreement triple for 42 judges on two community-built Indic benchmarks covering four domains and eight languages. Every comparison is reference-matched: a judge and a held-out rater are scored against the same two-rater mean, since an averaged reference otherwise flatters judges by several degrees. On subjective rubrics, judges agree with one another as much as humans do yet reach only 58–66% of human agreement: consensus on an axis humans do not weight. On the one rubric with a verifiable answer, most of that gap closes. Ensembles converge on the judges' shared axis rather than the human one, and training widens scores without rotating them. Inter-judge agreement is evidence of human alignment only after this check passes.[1]

## 1 Introduction

LLMs are now the default raters for open-ended text (Zheng et al., 2023; Chiang and Lee, 2023; Dubois et al., 2023). They track people closely when a rubric has a checkable answer, and far less closely on the subjective, multilingual, culturally situated judgements that dominate real deployments (Wang et al., 2024; Liu et al., 2023; Hada et al., 2024; Watts et al., 2024). Across meta-evaluations, one lopsided pattern recurs: judges agree strongly with one another while agreeing only weakly with people (Hada et al., 2024; Watts et al., 2024). How that pattern is read decides evaluation practice. If judges agree because they capture quality, ensembling them is sound and human annotation can be scaled back; if they agree because they share a blind spot, ensembling amplifies the error exactly where human judgement is scarcest. Agreement statistics cannot separate the two readings, because both produce the same numbers (Calderon et al., 2025).

We separate them by measuring the *shape* of the disagreement rather than its size. Suppose human judgement varies along many independent aspects of quality, while each judge's scores move along a few, largely shared directions (Schütze, 1992; Arora et al., 2018; Raghu et al., 2017). Judges built on overlapping pre-training corpora and similar post-training objectives (Ouyang et al., 2022; Bai et al., 2022) are exactly where this should occur, and it yields a prediction that agreement scores cannot test: judge and human scores should sit at a wide *angle* in score space, and averaging more judges should not close it. Throughout, "subspace," "axis," and "projection" name patterns in the scores we observe, never computation inside a model.

We test this prediction on two community-built Indic benchmarks: REAL-WORLD CULTURAL QA (Watts et al., 2024), with eight Indic languages, three domains, and three independent raters per item, and COMMUNITY-HEALTH (Hamna et al., 2026; Bhat et al., 2025), judged by 14 base LLMs and 27 trained variants of three open families (Brown et al., 2020; Rafailov et al., 2023). One measurement choice carries much of the weight. A judge compared against the average of several raters has an easier target than a rater compared against a single colleague, and the difference is worth several degrees; wherever we compare a judge with a person, we therefore hold out one rater and score *both* against the remaining two. Under this matched comparison the prediction holds. On every subjective rubric, judges agree with each other about as much as people do while reaching only 58–66% of that agreement with people, sitting 9–11° outside the human floor and using well under the human score range. The one rubric with a verifiable answer closes most of the gap. Training widens a judge's scores but barely rotates them (Fig. 1).

[1] Code and resources are publicly available.

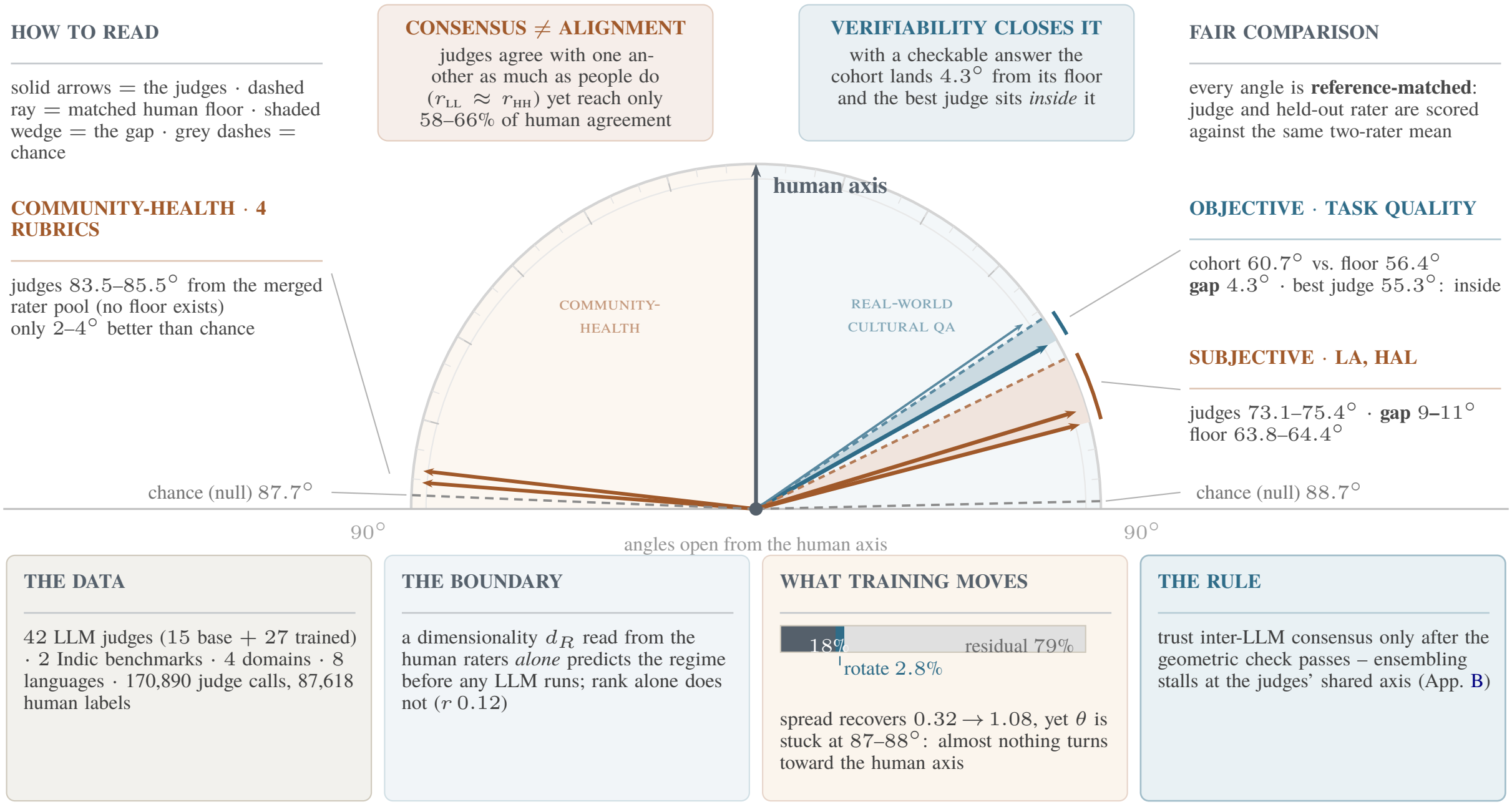

Figure 1: Angles open from the vertical human axis: solid arrows mark where LLM judges sit, dashed rays their *matched* human floor, and the shaded wedge between them, bracketed at the rim, is the gap this paper measures. On the one rubric with a checkable answer (teal) the cohort lands 4.3° from its floor and the best judge inside it; on the subjective rubrics (terracotta) the wedge opens to 9–11° even though judges agree with one another as much as people do. On COMMUNITY-HEALTH (left half) a single merged rater pool leaves no floor to draw, and the judges land only 2–4° from chance. The plates below the baseline carry the data, the boundary result, what training moves, and the rule. All quantities are observed score covariance, never model internals.

**Contributions.** This paper contributes: (1) a reporting frame that separates judge–judge, judge–human, and human–human agreement on a matched reference, with the geometric account scoped as a diagnostic over observed scores (Sec. 3); (2) four score-space diagnostics, validated by permutation, FDR, ordinal, human-reference, self-preference, and split-half controls that the collapse pattern survives (Sec. 5.4); (3) a falsifiable boundary: a dimensionality computed from human raters alone, before any LLM is run, predicts which regime a rubric falls in, with rank necessary and orientation decisive; (4) an anatomy of repair, showing that training restores spread rather than orientation and that ensembling converges on the judges' shared axis; and (5) a complete release that reproduces every number in this paper with one command (App. O). A gap to human judgement survives every training stage we tried, so multi-perspective human rating remains necessary.

## 2 Related Work

LLM-as-judge was adopted because it tracks humans on open-ended generation (Chiang and Lee, 2023; Zheng et al., 2023; Kocmi and Federmann, 2023; Dubois et al., 2023), and the reliability literature documented where it does not: fluency and verbosity bias, self-preference, and degradation on non-English and culturally situated content (Wang et al., 2024; Zheng et al., 2023; Hada et al., 2024; Watts et al., 2024; Hershcovich et al., 2022). Calderon et al. (2025) argue that agreement statistics alone cannot separate a reliable judge from a plausible one. Our alternative is diagnostic: rather than ask how large the LLM–human gap is, we ask what *shape* it has, and show the shape predicts when consensus may be trusted.

Consensus is where this matters most. Ensemble and panel protocols treat inter-judge agreement as reliability (Zheng et al., 2023), assuming averaging cancels idiosyncratic bias. That fails if judges share the bias: models trained on overlapping corpora with similar post-training objectives (Ouyang et al., 2022; Bai et al., 2022) can covary strongly while jointly missing a human-weighted direction. We make that failure measurable, and show ensembling converges on the cohort's own axis (Sec. 5.6).

Our instruments come from representational analysis: principal angles and canonical correlation (Björck and Golub, 1973; Knyazev and Argentati, 2002; Raghu et al., 2017), applied to *scores* rather than activations, which keeps the analysis model-agnostic and usable on closed APIs at the cost of a diagnostic rather than mechanistic lens (Sec. 3).

The setting is Indic NLP, a testbed for cross-lingual judge reliability (Watts et al., 2024; Hada et al., 2024). Our benchmarks supply complementary halves of what the diagnostics need: REAL-WORLD CULTURAL QA (Watts et al., 2024) gives three independent raters per item, which makes a human floor measurable at all, while COMMUNITY-HEALTH (Hamna et al., 2026; Bhat et al., 2025) gives judge breadth. We evaluate four standard repairs (Brown et al., 2020; Ouyang et al., 2022; Rafailov et al., 2023; Hu et al., 2022; Platt, 1999; Guo et al., 2017); new here is the decomposition (Sec. 5.8) separating amplification of a judge's ranking from rotation toward humans.

## 3 A Geometric Lens on Judge Score Matrices

This section defines what the paper measures. It is a *diagnostic lens over observed scores*: "projection," "subspace," and "axis" refer to structure in observed score covariance, never to internal LLM computation. Every analysis operates on per-judge score matrices $\mathbf{S}_{\mathcal{J}} \in \mathbb{R}^{n \times m}$ ($n$ items, $m$ rubrics; columns mean-centered). Two constructions recur: *per-rubric*, where a judge column $\mathbf{s}$ meets a reference column $\mathbf{h}$ and the angle reduces exactly to $\theta = \arccos |r(\mathbf{s}, \mathbf{h})|$ for Pearson $r$; and *stacked*, the largest principal angle (Björck and Golub, 1973; Knyazev and Argentati, 2002) between the 95%-energy singular bases of the full matrices. App. D expands this section with the idealised picture, the worked identity, and the boundary case analysis; App. L gives procedures, preprocessing/cutoff grids, and the permutation-null baseline; App. N reconciles the constructions numerically.

We track four diagnostics.

(i) *Spread ratio* $\sigma_{\mathcal{J}}/\sigma_H$: judge over human-reference score SD, per rubric.

(ii) *Effective rank*

$$r_{95}^{\mathcal{J}} = \min\{k : \textstyle\sum_{i \le k} \lambda_i / \sum_i \lambda_i \ge 0.95\}$$

(Jolliffe and Cadima, 2016): how many independent directions the judge's scores use.

(iii) *Principal angle* $\theta(\mathcal{S}_{\mathcal{J}}, \mathcal{S}_H)$, always read against a *reference-matched* inter-human floor (where multiple raters exist; Sec. 5) *and* a permutation null: near-orthogonality is generic in high dimension, so an angle is informative only relative to that null.

(iv) *The agreement triple* $\overline{r}_{\text{LL}}, \overline{r}_{\text{LH}}, \overline{r}_{\text{HH}}$, reported together, never as one aggregate.

**Boundary: rank necessary, orientation decisive.** Let $d_R$ measure the dimensionality of *human rater disagreement* under rubric $R$, computed *from human raters alone* as the participation ratio of the centered items×raters matrix,

$$d_R = \Big(\textstyle\sum_i \lambda_i\Big)^2 \Big/ \textstyle\sum_i \lambda_i^2, \qquad (1)$$

with $\lambda_i$ the squared singular values. Note $d_R \le R$ for $R$ raters, so with three raters it is capped at 3 and reports disagreement dimensionality at that panel size, not the intrinsic dimensionality of human quality. A cohort with $k_{\mathcal{J}} < d_R$ cannot represent the human variation, so $k_{\mathcal{J}} \ge d_R$ is necessary, but not sufficient, since a judge may spend its rank on directions humans ignore. The decisive quantity is orientation. For a single direction the relationship is an identity rather than an analogy: since $\cos\theta = |r|$,

$$1 - \cos^2\theta(\mathbf{s}, \mathbf{h}) \;=\; 1 - r^2, \qquad (2)$$

which is exactly the fraction of human score variance that direction leaves unexplained; App. D works the numbers through and connects them to the ridge estimates. A stronger, rank-only gate is falsified by our own measurements ($r_{95}$ does not predict alignment, $r = 0.12$; Sec. 5.4). A judge is *rank-matched* on a rubric when its angle beats the permutation null and reaches the matched inter-human floor with $r_{\text{LH}} \approx r_{\text{HH}}$.

**Falsifiable predictions.** Tested in Sec. 5: **P1** judges compress spread on subjective rubrics; **P2** judge effective rank sits at or below the human reference's; **P3** on the three-rater benchmark, judge–human angles exceed the matched inter-human floor on subjective rubrics and approach it on the objective one, significantly below the permutation null in both regimes; **P4** $\overline{r}_{\text{LL}} > \overline{r}_{\text{LH}}$ on subjective rubrics via a dominant shared score axis, measurable by cohort PCA; **P5** $d_R$ from humans alone separates the regimes; **P6** rank alone does *not* predict alignment;

orientation does. The lens predicts *which* rubrics fail and the *shape* of failure, not which intervention repairs it (Sec. 5.7).

## 4 Experimental Setup

The study spans 170,890 LLM-judge calls ($\approx$474,310 rubric-level scores and preference labels) over two community-built Indic benchmarks covering four domains, eight Indic languages, and 42 distinct LLM judges (15 base models plus 27 fine-tuned, preference-optimised, few-shot, or calibrated variants), evaluated against four human reference pools ($\approx$87,618 human labels; App. E). Every roster, per-cell count, and event total is regenerated from the released data by one script (App. O).

**Datasets and languages.** COMMUNITY-HEALTH (Samiksha (Hamna et al., 2026; Bhat et al., 2025); **CH**) is a medical-advisory benchmark in Hindi, Kannada, and Malayalam: 810 questions (270/language), each answered by three generator LLMs (= 2,430 scored QA pairs) with paired standalone and comparative human labels; all geometric/training analyses use its 405-item core ($45\times3\times3$, the subset with complete topic metadata), comparative agreement the full pair set. REAL-WORLD CULTURAL QA (Pariksha (Watts et al., 2024); **DLQ**, daily-life QA) contributes **health & wellness**, **finance**, and **everyday life** with *three independent raters per item*; we retain the 1,182 items all three raters completed (300 Hindi down to 60 Odia; Table 8), from 27 answer generators. The eight analysed languages (Bengali, Hindi, Kannada, Malayalam, Marathi, Odia, Tamil, Telugu) span Indo-Aryan and Dravidian families; DLQ's comparative split (which adds Gujarati and Punjabi) is excluded from analysis (a single LLM judge with a label-orientation artefact; App. M) and counted only in event totals.

**Judges.** 15 base LLMs cover closed-API frontier, open multilingual, and Indic-focused families (open weights 12B–671B): 14 judge CH, 5 judge DLQ (deepseek-r1, gpt-4, grok-4, mistral-large-3, llama-3.3-70b), four anchoring both. Three Indic-capable open families (sarvam-m-24b, gemma-3-27b, aya-expanse-32b) are exercised on CH under six techniques: few-shot (Brown et al., 2020), LoRA SFT (Hu et al., 2022; Ouyang et al., 2022), DPO (Rafailov et al., 2023), SFT+DPO, demonstration-augmented SFT (demos baked into training prompts), and demonstration-anchored calibration (a frozen trained checkpoint scored with $k$ human-rated demos in context); SFT/DPO train on held-out CH-source labels with *zero question overlap* with the evaluation set. Rosters, recipes, hyper-parameters: Tables 10–15; API versions and call dates: App. O.

Every QA pair is scored under two paradigms. **Standalone**: per-rubric ordinal scores over four subjective CH rubrics (*clarity*, *helpfulness*, *accuracy*, *completeness*) plus three DLQ rubrics: *task quality* (the single objective rubric: verifiable task constraints), *linguistic acceptability*, and *hallucinations* (Chiang and Lee, 2023; Watts et al., 2024). **Comparative**: judges pick the better of two answers. The paradigms separate *calibration* from *rank-ordering* failures (Zheng et al., 2023; Liu et al., 2023); definitions and verbatim prompts in App. G, I.

**Measurement and inference.** Per (judge, rubric, language, domain, eval type) slice we compute the Sec. 3 diagnostics: the agreement triple (quadratic-weighted $\kappa$ in App. M); Cohen's $\kappa$ (Cohen, 1960) on comparative panels; $\sigma_{\mathcal{J}}/\sigma_H$; $r_{95}$; and $\theta$ against the matched human floor and permutation null (App. L). The training analysis uses a paired item bootstrap (Efron, 1979), $B = 1000$ over the 405 core items with rows reused across (family, stage); with $n = 3$ trainable families the smallest one-sided Wilcoxon (Wilcoxon, 1945) $p$ is 0.125, so Wilcoxon is only a directional summary.

## 5 Empirical Analysis

Four analyses plus controls test the geometric account: agreement, boundary, geometry, controls (testing P1–P6; Sec. 5.4), and training.

**One construction per comparison.** Two conventions govern every number below, since mixing them is the easiest way to manufacture a result. *(i)* All three legs of an agreement triple are **pairwise** (rater vs. rater): correlating a judge with the *mean* of three raters gives it a less noisy target than the other two legs face, which on task quality alone moves $r_{\mathrm{LH}}$ from 0.418 to 0.519 and would lift judges above the human floor. *(ii)* Every angle compared with a human floor is **reference-matched**: one rater is held out and the mean of the other two is the single reference that this rater *and* each judge are scored against. App. N reconciles the constructions.

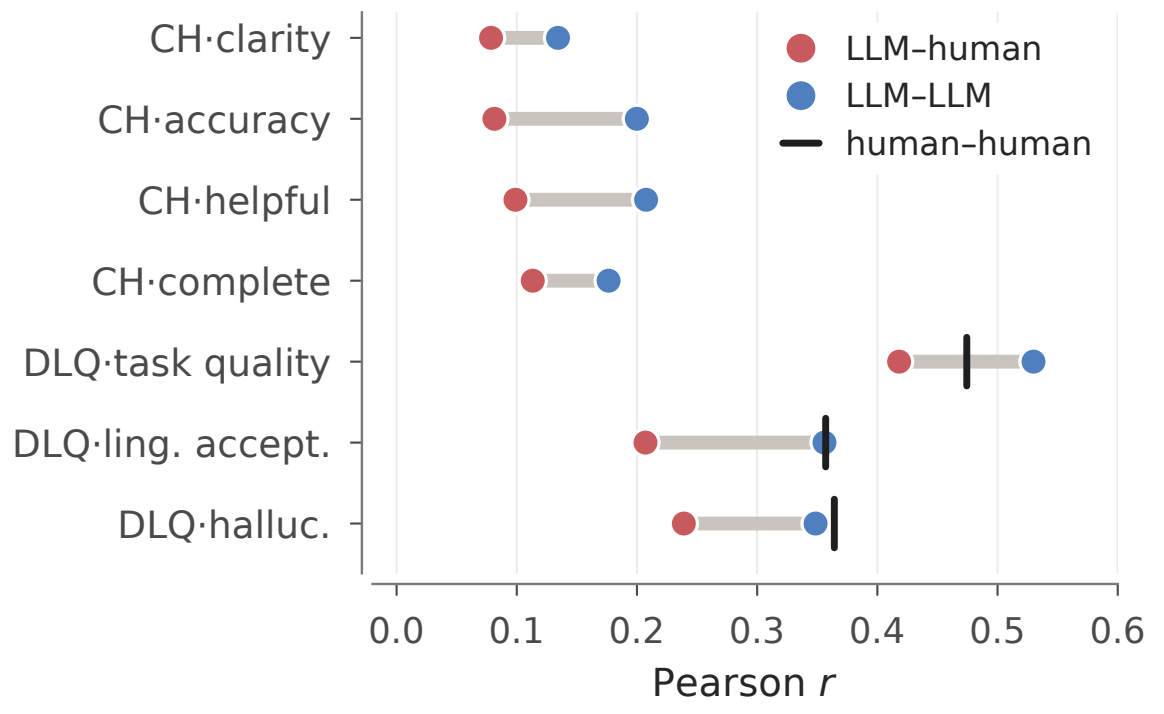

Figure 2: Segment length = the LL−LH gap, all legs pairwise. Extended caption in Table 6.

## 5.1 Inter-LLM consensus versus LLM–human agreement

On REAL-WORLD CULTURAL QA's subjective rubrics the three legs separate cleanly: judges agree with each other about as much as humans agree with each other ($r_{\text{LL}}$ 0.356 vs. $r_{\text{HH}}$ 0.357 on linguistic acceptability; 0.349 vs. 0.364 on hallucinations) while agreeing with humans far less (0.207, 0.239, i.e. 58% and 66% of the inter-human level). Matching inter-human *consensus* without matching inter-human *agreement* is the paper's central observation: consensus is reached on a different axis. The collapse is strongest on COMMUNITY-HEALTH (41–64% of inter-LLM across four rubrics), and its comparative split sharpens it in preference space: LLM–LLM $\kappa = 0.445$ vs. LLM–human $\kappa = 0.128$, a 3.5× gap (base judges, 2,430 pairs, three-class $\kappa$). Table 1A gives the full triple and the matched angles; Fig. 2 maps every cell's gap.

## 5.2 The boundary is rubric-conditional

The strongest reading of consensual blindness, that LLM judges are generically unreliable, is falsified by REAL-WORLD CULTURAL QA: with the same items, the same five judges (deepseek-r1, gpt-4, grok-4, mistral-large-3, llama-3.3-70b), and the same three humans, task quality lifts LLM–human agreement to 0.418, or 88% of the inter-human level (0.474), against 58% and 66% on the two subjective rubrics. The angle tells the same story with the reference matched: judges sit 60.7° from the two-rater reference where a held-out human sits 56.4°, a gap of 4.3°, against 11.0° and 9.3° on the subjective rubrics. The best single judge (55.3°) is inside the human floor on task quality; on both subjective rubrics the best judge (67.3°, 70.8°) remains well outside it. Figure 3 shows this item by item: the subjective clouds stay flat against the identity line, task quality tracks it.

| cohort | n | $\sigma_{\mathcal{J}}/\sigma_H$ | $\theta^{\circ}$ | $\theta^{\circ}_{\text{HH}}$ | $r$ | $r_{\text{ref}}$ | $r_{\text{HH}}$ | $\Delta$ |
|---|---|---|---|---|---|---|---|---|
| **(A) Rubric×dataset** *matched $\theta$; pairwise $r$; $r_{ref}$: inter-LLM* | | | | | | | | |
| CH subj. mean | 14 | 0.44 | 84.4 | – | 0.093 | 0.180 | – | +0.087 |
| DLQ LA | 5 | 0.734 | 75.4 | 64.4 | 0.207 | 0.356 | 0.357 | +0.149 |
| DLQ HAL | 5 | 0.784 | 73.1 | 63.8 | 0.239 | 0.349 | 0.364 | +0.110 |
| DLQ TQ (obj.) | 5 | 0.687 | 60.7 | 56.4 | 0.418 | 0.530 | 0.474 | +0.112 |
| **(B) Base-judge strata on CH** *stacked; $r_{ref}$: pool mean* | | | | | | | | |
| FE | 4 | 0.635 | 87.5 | – | 0.131 | 0.093 | – | +0.039 |
| OXL | 4 | 0.368 | 88.5 | – | 0.070 | 0.093 | – | −0.023 |
| OM | 4 | 0.222 | 88.3 | – | 0.082 | 0.093 | – | −0.011 |
| IF | 2 | 0.641 | 89.6 | – | 0.083 | 0.093 | – | −0.010 |
| **(C) Training stage on CH** *stacked; $r_{ref}$: BASE* | | | | | | | | |
| BASE | 3 | 0.319 | 87.7 | – | 0.081 | 0.081 | – | 0.000 |
| FS | 3 | 0.624 | 88.3 | – | 0.122 | 0.081 | – | +0.041 |
| FT | 3 | 1.074 | 87.1 | – | 0.129 | 0.081 | – | +0.048 |
| FT+PO | 3 | 1.078 | 87.1 | – | 0.132 | 0.081 | – | +0.051 |
| AT[4] | 1 | 0.649 | 88.0 | – | 0.177 | 0.081 | – | +0.096 |
| CAL[4] | 1 | 0.633 | 88.1 | – | 0.184 | 0.081 | – | +0.103 |

Table 1: **Consolidated results grid.** $\Delta = r_{\text{ref}} - r$ in panel A (how far LLM–human sits below inter-LLM) and $r - r_{\text{ref}}$ in B–C. Panel A: all three $r$ legs pairwise; $\theta$ and $\theta_{\text{HH}}$ reference-matched, so the judge and the held-out human face the same two-rater target (DLQ only: CH merges its annotators into one reference, so no floor exists there and its $\theta$ is against that single reference). Panels B–C: per-judge stacked angles (C: bootstrap stage means), which run higher than panel A by construction (Table 4; App. N). [4]: lifts all four CH rubrics over BASE (unmarked trained rows do not; base judges take no marker). Shaded cells mark the decisive numbers: sand = the worst case in a panel, mist = the relief case. Key, codes, strata: App. H.1.

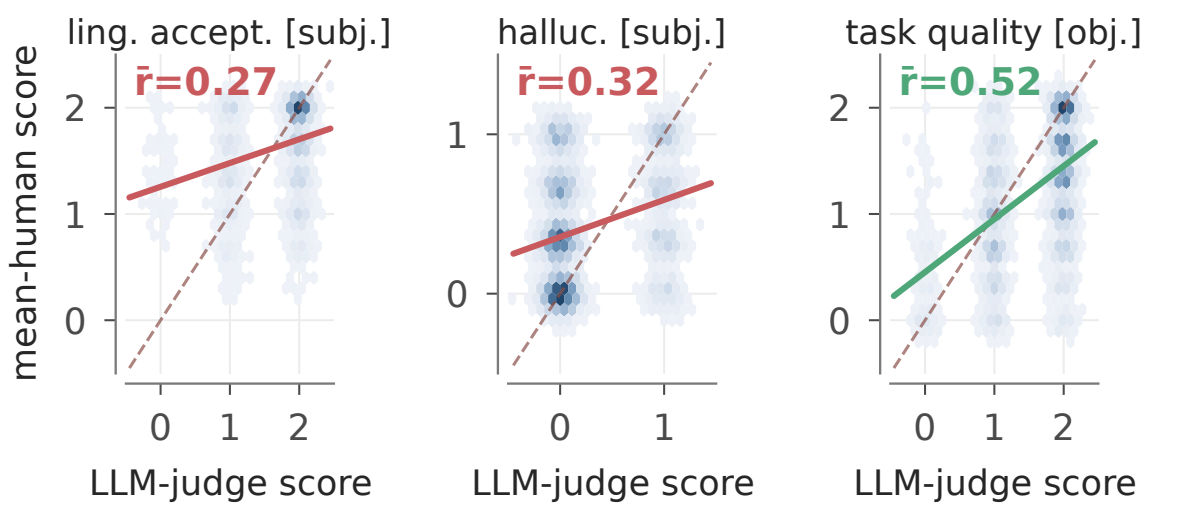

Figure 3: Three-rater mean human score against LLM-judge score for the five DLQ base judges, one panel per rubric (hexbin density, dashed identity line, solid least-squares fit). Axes follow each rubric's own ordinal scale ($\{0, 1, 2\}$; hallucinations is binary). $\bar{r}$ is the judge-vs-mean-human Pearson, the quantity these panels plot; the pairwise leg comparable with inter-human agreement is lower (0.418/0.207/0.239) and is what Table 1A reports. The two subjective rubrics stay flat against the identity line; task quality tracks it.

The boundary is a large change of degree, not a crossing: judges close most but not all of the gap where the rubric is verifiable. Scoring judges against the three-rater mean while scoring humans against each other puts task quality *above* the human floor, but that ordering is an artefact of the mismatched reference and does not survive matching.

We do not claim that LLM-as-judge fails in general: subjective rubrics with high rater-disagreement dimensionality expose a mismatch between the judge and human subspaces, whereas objective rubrics can be rank-matched. Items illustrating both regimes (factual recall and shared cultural knowledge on one side; healthcare access, affordability, overmedicalisation, and emotional framing on the other) are in Tables 26–27 (App. P).

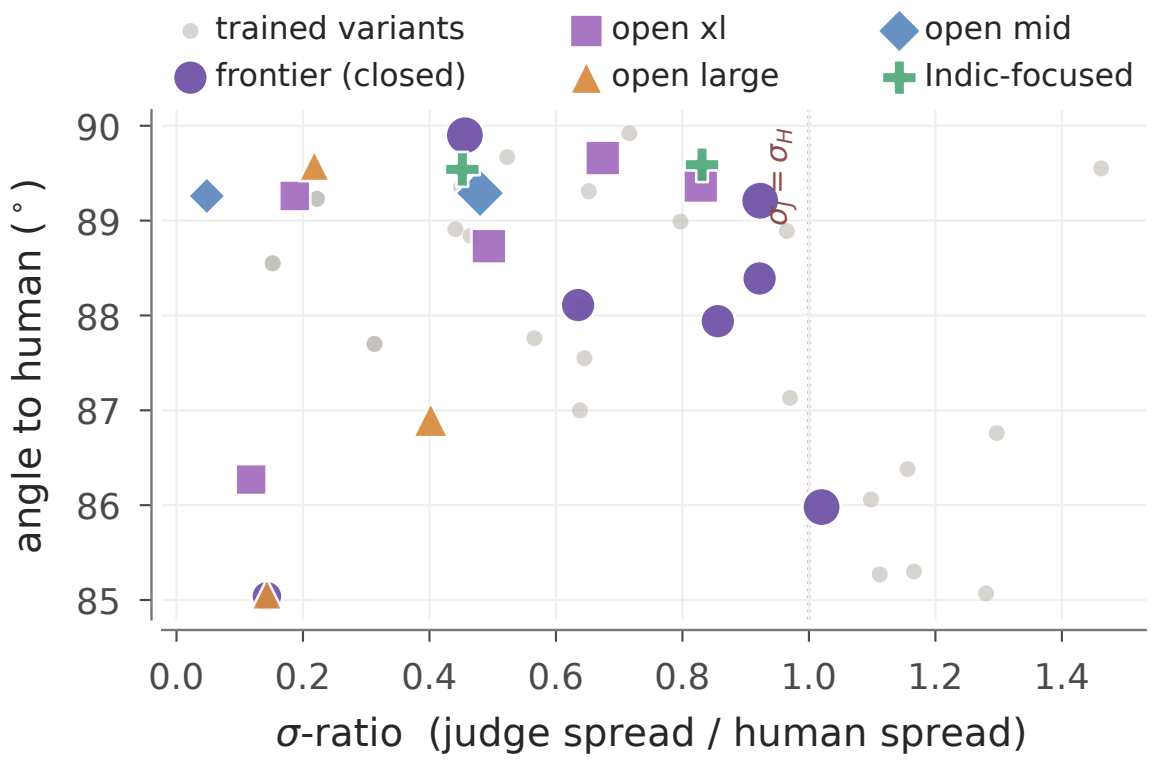

Figure 4: All 40 judges in ($\sigma$-ratio, angle) space; position and spread, not a ranking. Extended caption in Table 6.

### 5.3 Geometric signature of the disagreement

The failure is not only a correlation gap; it has a characteristic geometry (Table 1): compressed spread, low rank, high angle, large unrecoverable human variance (Sec. 5.4). On CH the median base judge uses under half the human spread ($\sigma$-ratio 0.454) at a stacked angle of 89.3°. On DLQ, with the reference matched, base judges sit 84.3–88.9° from the two-rater reference while a held-out human sits 73.9–78.1° from it: a gap of roughly 10° that Pearson alone does not expose. Rank confirms P2: human references use the full basis ($r_{95} = 4$ CH, 3 DLQ); 11 of 14 CH base judges sit at $r_{95} \leq 3$. Angles are comparable only within one construction (Table 4; App. N).

Training reliably moves judges rightward (spread) but not downward (angle); Fig. 4 plots every judge in that space.

Task quality does not merely raise Pearson, it *rotates* the judge vectors: 4.3° from the floor against 9.3–11.0° on the subjective rubrics, even though spread stays compressed ($\sigma$-ratio 0.687; $\sigma_H$ is the mean single-rater SD throughout, matching CH's single merged reference). Alignment and spread calibration are separable: on the objective rubric judges point close to the right way but under-use the range (Fig. 5).

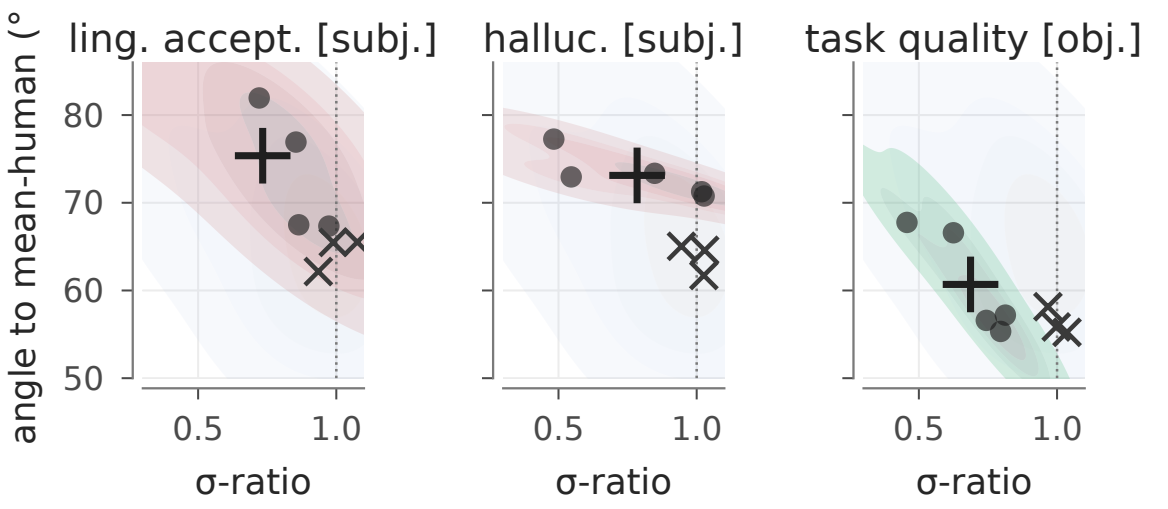

Figure 5: One panel per rubric, judges and held-out humans on the matched two-rater reference. Extended caption in Table 6.

| Control | Outcome |
|---|---|
| Permutation null | DLQ per-rubric null 88.6–88.7°; every judge×rubric angle below *all* 500 shuffles ($p < 0.002$ each) |
| CH judge-level + FDR | CH stacked null 88.8°; 11/40 judges below at $p < .05$ and 6 survive Benjamini–Hochberg at $q < .05$ (20,000 shuffles) |
| Ordinal / scale | Spearman = Pearson within 0.02; remap onto human marginal moves $\theta$ by $\leq 0.13°$ (DLQ) / $\leq 0.31°$ (CH) |
| Self-preference | judges from a generator's own family rate it no differently than the rest of the cohort does ($\lvert\Delta\rvert \leq 0.035$ SD, all $p \geq 0.58$) |
| Split-half | item-half angles agree to $\leq 2.5°$; DLQ judge ordering across halves $r = 0.74$–$0.97$; CH judge ordering does *not* replicate |
| Unrecoverable var. | ridge CV: 55% (TQ), ≈ 80% (DLQ subj.), ≈ 100% (CH) of human variance unrecoverable from the judge cohort |
| $d_R$ separator (P5) | human-only dimensionality splits objective (≈ 2.1) from subjective (≈ 2.4) cells before any LLM runs; no graded trend within either group |

Table 2: **Controls summary.** Procedures, per-rubric values, judge-level permutation results with $q$ values, and caveats in App. L, M.

### 5.4 Controls: is the geometry real?

Control families, all on released scores with no new inference, test whether the geometry is an artefact and test P1–P6 (Table 2; procedures in App. L–M).

Against a permutation null (500 shuffles/judge; nulls 88.6–88.7°), *every* DLQ judge×rubric angle beats all shuffles ($p < 0.002$, all 15 tests), and the weakest survives Bonferroni across all 15: task quality at 60.7° sits *significantly below the permutation null*, and the subjective elevations above the floor are consistent offsets (P3). On CH the cohort sits nearest the null (88.1° vs. 88.8°), the strongest collapse we report, yet individual judges still depart from it: at 20,000 shuffles, 11 of 40 judges are below the null at $p < .05$ (2 expected by chance) and 6 survive Benjamini–Hochberg (Benjamini and Hochberg, 1995) at $q < .05$, including two of the four frontier closed judges (gpt-5.5, gpt-4o), llama-3.3-70b, and three fine-tuned variants of the Indic-capable open families (Table 19).

That is each judge against *chance*, not against *each other*. The permutation is paired, holding a judge's matrix fixed while shuffling the human rows, so it removes item-sampling variance; an item bootstrap does not, and gives each CH angle a 95% interval 6.3° wide against a between-judge SD of only 1.6°. We therefore rank no CH judges by angle anywhere, and CH comparisons use stacked

Pearson (Sec. 5.7).

Nor is compression merely cautious scale use: Spearman tracks Pearson to within 0.02 per rubric, and monotone remapping onto the human marginal moves angles by $\leq 0.13°$ (DLQ) / $\leq 0.31°$ (CH), so judges occupy a *different direction*, not a narrower stretch. We also report a magnitude rather than an all-or-nothing "null-space" claim: cross-validated ridge reconstruction from the *entire* cohort leaves 55% of human variance unrecoverable on task quality, $\approx 80\%$ on subjective DLQ, essentially 100% on CH (CV $R^2 \leq 0$).

The regimes also separate before any LLM is run (P5). The participation ratio $d_R$ (Eq. 1), computed from the three human raters alone, tracks the LLM–human angle across the nine DLQ rubric×domain cells ($\rho = 0.68$, $p = 0.042$, $R^2 = 0.83$, leave-one-cell-out stable). The relation is *between* regimes, not graded within them: objective cells cluster at $d_R \approx 2.1$ and $\approx 60°$, subjective at $\approx 2.4$ and $\approx 74°$, while inside the six subjective cells the correlation vanishes ($\rho = -0.03$). We read $d_R$ as a two-regime separator, not a dose–response curve, and quote no slope; with three raters it is capped at 3, measuring *rater disagreement* at this panel size (see Limitations). The supported claim is the useful one: *a quantity computable from humans alone tells you which regime a rubric is in.*

Self-preference is not the driver either. Two of CH's three answer generators (sarvam-m, qwen3) also serve as judges, and sarvam-m supplies the headline calibrated judge, so family favouritism could manufacture the training result. Within a (language, question) triple, a judge's standardised deviation from its triple mean, minus the human's and then the rest of the cohort's, isolates family preference from item difficulty and real quality differences. No family shows an effect ($|\Delta| \leq 0.035$ SD; permutation $p \geq 0.58$; App. M.8).

Finally, rank is necessary but orientation is decisive (P6): across the 40 CH judges effective rank does not predict alignment ($r = 0.12$; $r_{95}$ spans 1–4, angles 85.0–89.9°). With task quality still compressed ($\sigma$-ratio 0.69) yet rotated to within 4.3° of the floor, orientation, not rank or spread, carries alignment (Eq. 2).

### 5.5 Breadth: domains and languages

The remaining diagnostics use DLQ, the only setup with per-cell human floors. **Across domains** the boundary reproduces in all three (Fig. 6): on task quality the LLM mean sits 60.6–63.3° against a matched floor of 57.7–58.2°, with LH Pearson at 85–92% of H–H; on the subjective rubrics the gap widens to 8.3–14.3° and LH falls to 49–66%. $\sigma$-compression (0.67–0.82) is domain-invariant; Fig. 14 (App. H) breaks agreement out by language, domain, and rubric kind across all four domain datasets.

**Across the eight languages** the robust findings are compression and the agreement gap, not the angle. $\sigma$-compression is universal (0.43–0.76) and the LL−LH gap is positive in all eight, but its size ranges over two orders of magnitude, from +0.227 (Bengali) to +0.005 (Telugu, effectively closed). Per-language *angles* we do not read: at $n = 60$–300 per language the stacked angle saturates near 90° for judges and humans alike, and only 26 of 40 judge×language cells sit above their own matched floor (Fig. 12).

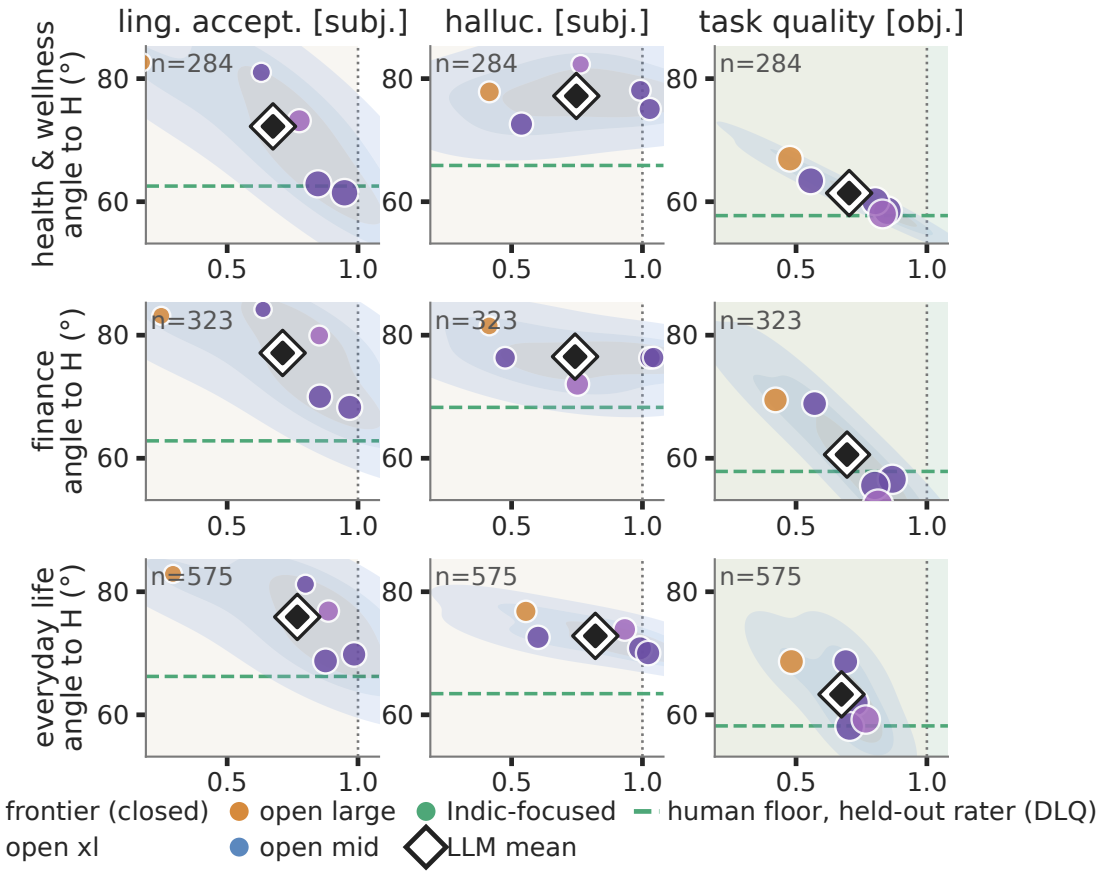

Figure 6: Rows = domains, columns = rubrics; each panel plots base judges in ($\sigma_{\mathcal{J}}/\sigma_H$ horizontally, angle vertically), both against the matched two-rater reference, with the dashed green line the held-out-rater floor and pale green marking the objective rubric. The objective column drops toward the floor in every domain; the two subjective columns sit clearly above it. CH geometry is in Table 1A and Fig. 2.

### 5.6 Subspace topology of the judge cohort

Do judges sit in the *same* direction, or split into sub-clusters equally far from humans? Z-scoring each judge's per-rubric scores over the 405 CH core items and taking cosines between the 1620-d item×rubric vectors measures pure orientation ($\overline{\cos}_{\text{LL}} = 0.16$ vs. $\overline{\cos}_{\text{LH}} = 0.08$; Fig. 7).

Three features are invisible to per-judge angle plots. *(i)* The tightest pairs do *not* follow model family: the three deepest cells are all cross-vendor, and the only same-family pair (qwen3, 0.50) ranks fourth against a 0.16 cohort mean (values in Table 6); scale and openness do not organise the map either, so the shared orientation is a cohort-wide property rather than siblings agreeing. *(ii)* The cohort splits into a tight frontier+open block (intra 0.27) and a looser mid-tier block (intra 0.11; across 0.08): a single **LL** aggregate averages these into a consensus no judge holds. *(iii)* The human row stays uniformly pale after Z-scoring, so the collapse belongs to the LLM-shared subspace, not to a scale or single-judge artefact.

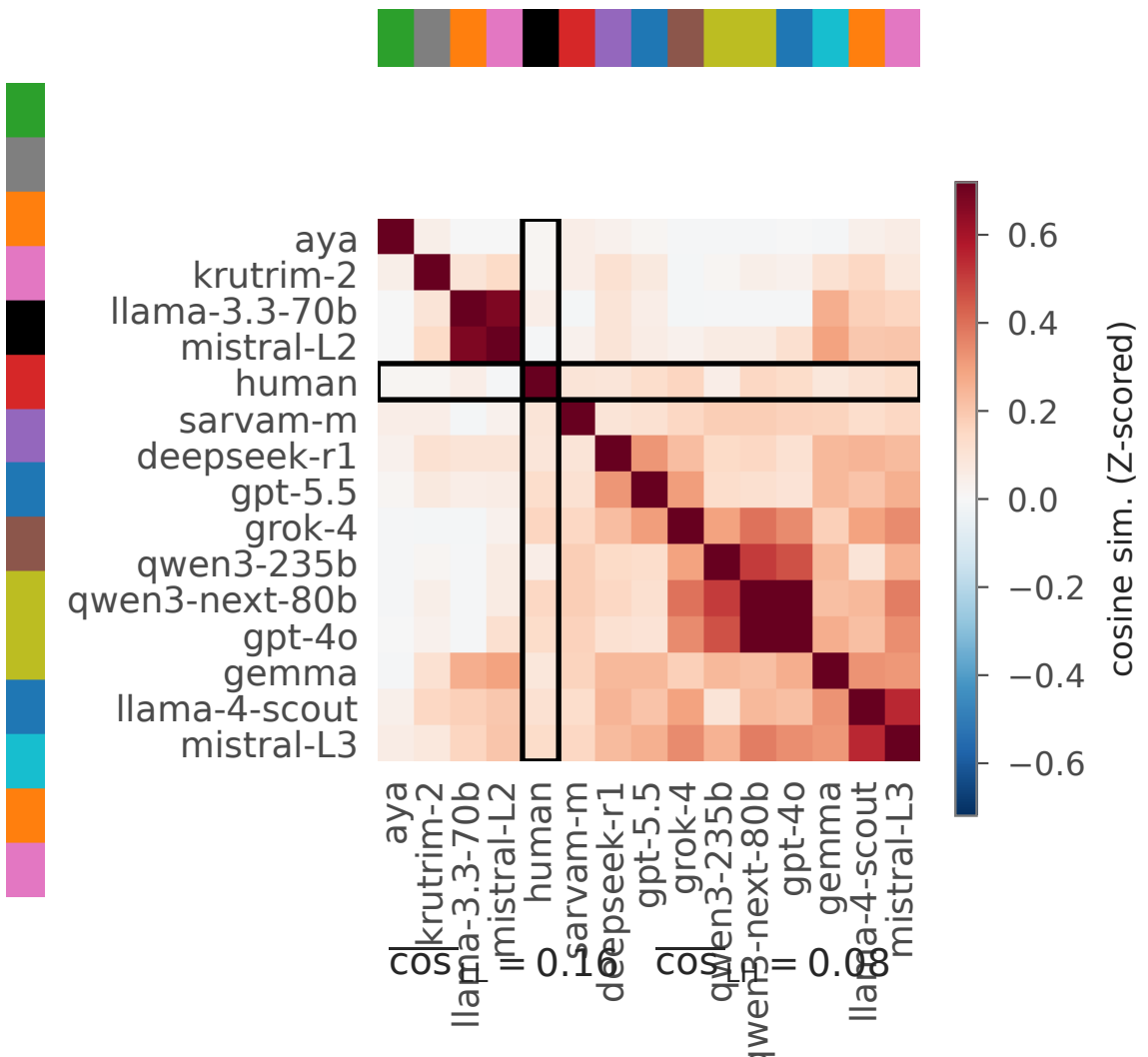

Figure 7: Ward-reordered cosines; the human row stays uniformly pale. Extended caption in Table 6.

| **Rubric** | $k$=1 | 2 | 3 | 4 | 5 | $k \to \infty$ | **floor** |
|---|---|---|---|---|---|---|---|
| task quality | 60.7 | 55.5 | 53.3 | 52.0 | 51.2 | 50.0 | 56.4 |
| ling. accept. | 75.4 | 71.2 | 69.4 | 68.3 | 67.6 | 66.5 | 64.4 |
| hallucinations | 73.1 | 69.1 | 67.2 | 66.2 | 65.5 | 64.6 | 63.8 |

Table 3: **Ensemble-size curve (DLQ, matched reference).** Mean angle° over all $\binom{5}{k}$ subsets of the five base judges, each scored against the same held-out two-rater mean that defines the floor. $k \to \infty$ sums the geometric tail of the decrements. Mist marks ensembles already inside their floor, sand a limit that stays outside it: on task quality two judges suffice, on both subjective rubrics no ensemble of any size reaches the floor. Caveats in App. M.10.

The shared axis is directly measurable (P4): the cohort's top three PCs explain 86.6–89.0% of judge variance on DLQ (66–82% CH), and the dominant axis sits 47.7° from humans on task quality but 64–66° on the DLQ subjective rubrics (79–86° on CH). Averaging moves toward that axis and stops there (Table 3): ensembling $k$ judges lowers the matched angle monotonically, but the decrements shrink by a near-constant factor (≈0.6), so the geometric tail bounds an unlimited ensemble of these judges outside the subjective floors. Ensembling buys variance reduction toward the shared axis, not a new direction (App. M.10).

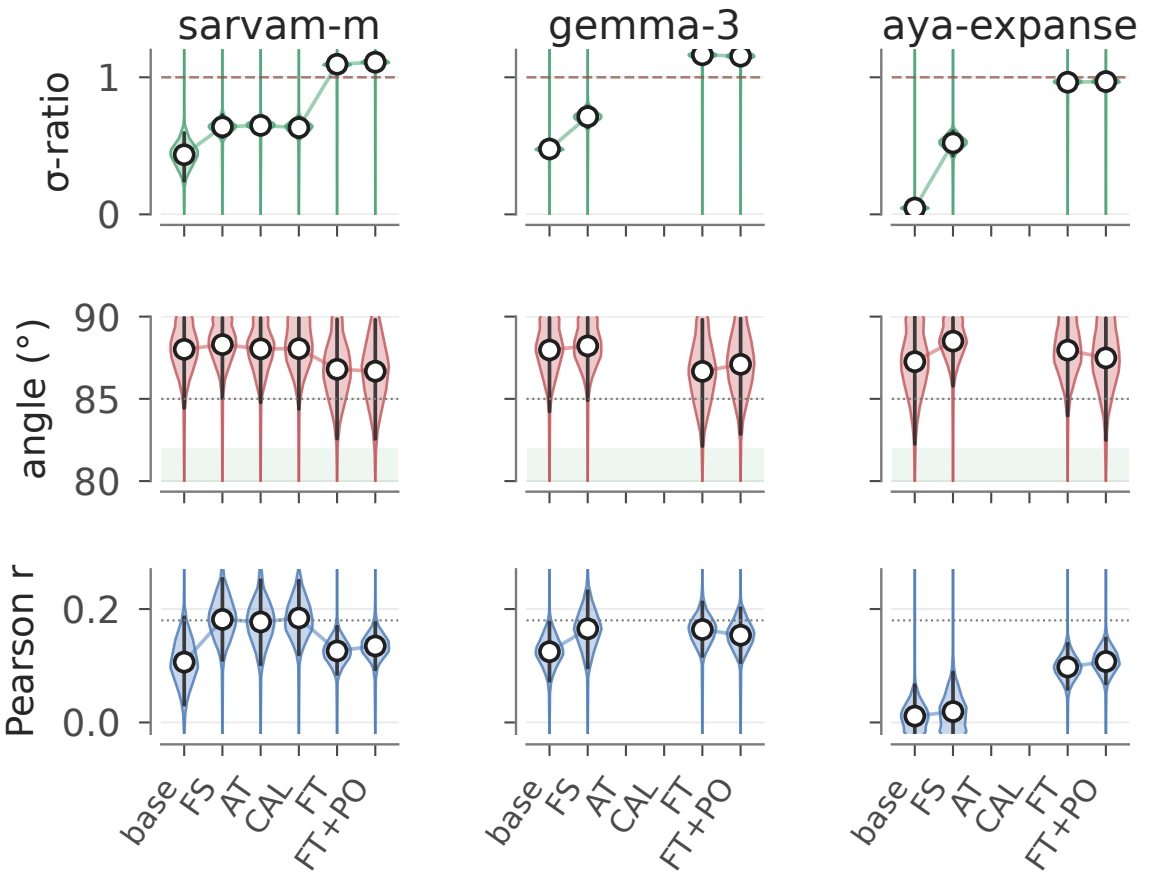

Figure 8: Bootstrap over $B = 1000$ resamples. Spread rises reliably; angle distributions overlap throughout. Extended caption in Table 6.

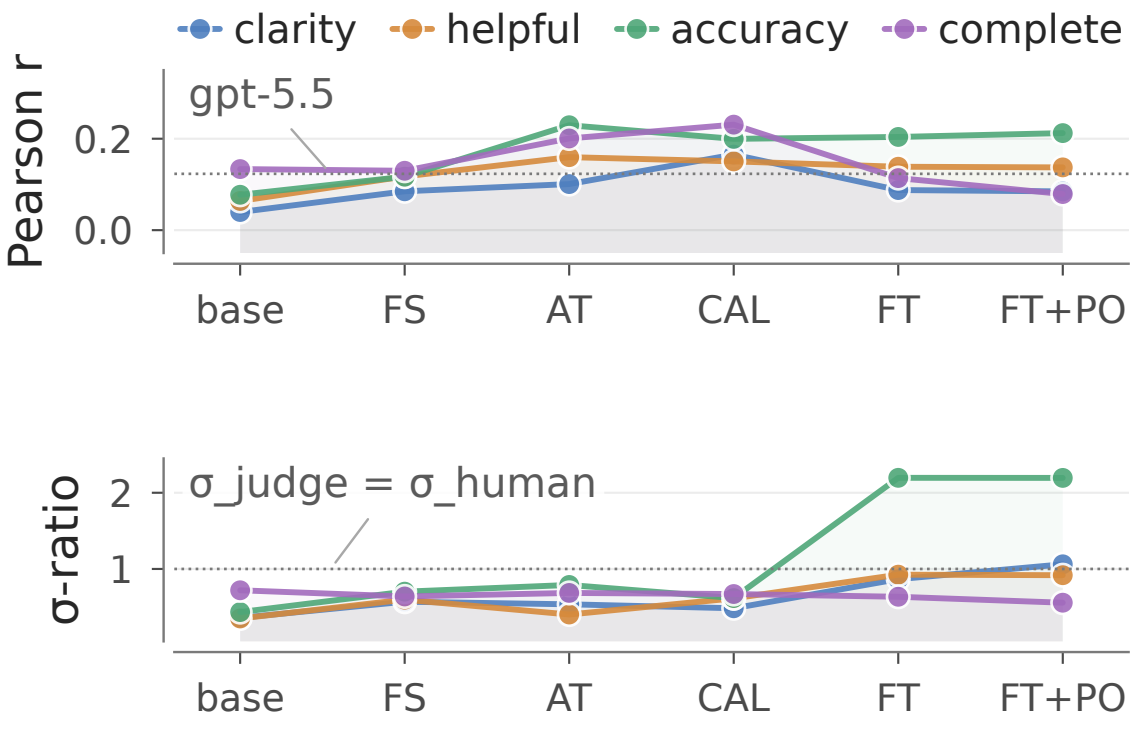

Figure 9: Only the demonstration-anchored stages raise every rubric. Extended caption in Table 6.

### 5.7 Training restores spread before orientation

Which part of the geometry can training change? On COMMUNITY-HEALTH we follow sarvam-m 24B, gemma-3 27B, and aya-expanse 32B through base, few-shot, fine-tune, fine-tune+pref-optim, and (sarvam-m only) demonstration-augmented SFT (AT) and demonstration-anchored calibration (CAL), picking the best in-stage judge per family by stacked Pearson with paired item-bootstrap CIs. $\sigma$-ratio rises strongly under FT and FT+PO ($0.32 \to 1.08$) but the angle barely moves: stage means stay at 87–88° with overlapping CIs (Table 1C; Fig. 8).

The aggregate masks per-rubric reversals (Fig. 9): FT/PO over-spreads accuracy ($\sigma \approx 2.2$) and degrades completeness ($0.134 \to 0.079$), and few-shot leaves completeness flat, so only the two *demonstration-anchored* stages lift all four rubrics. Both were run on sarvam-m alone, so this is a within-family ordering, not a cross-family comparison. Calibration puts the best 24B Indic judge at $r = 0.184$, above gpt-5.5 (0.123; Fig. 13) yet far

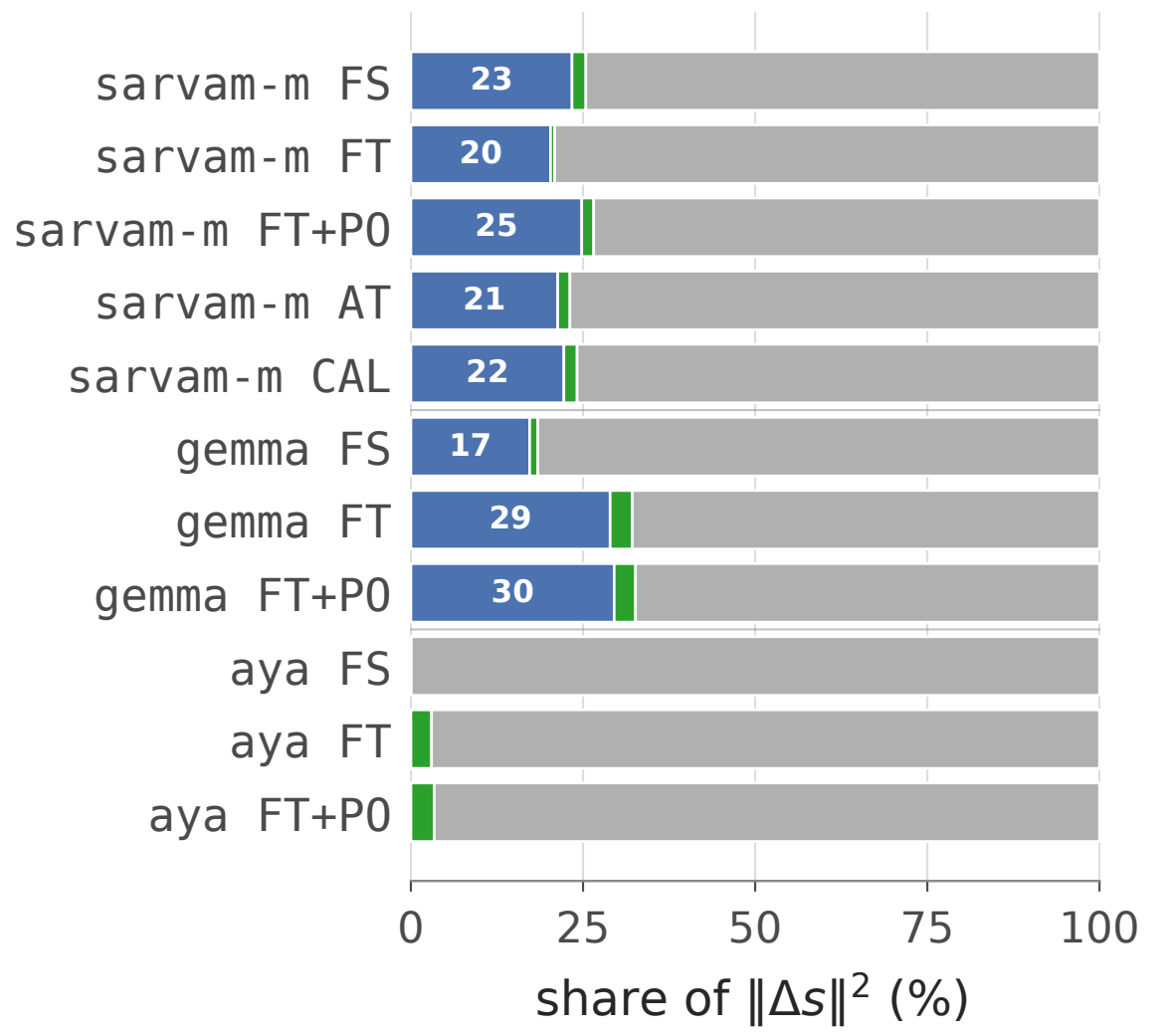

Figure 10: Each bar splits $\|\Delta s\|^2$ into stretch along the base axis (blue, labelled), rotation toward the human axis (green, labelled), and off-axis residual (grey). Rotation toward the human axis is $\leq$ 3.5% in every stage cell. Extended caption in Table 6.

below DLQ inter-human agreement (0.474).

### 5.8 Decomposing the training delta: stretch, rotate, residual

The bootstrap says what training *outputs*, not *where each unit of delta goes*. Per (family, stage) we decompose $\Delta s = s_{\text{stage}} - s_{\text{base}}$ over the 405 core items into orthogonal energies: **stretch** along the base axis, **rotate→human** (the only direction that can shrink the angle), and **off-axis residual** (Fig. 10).

Two findings give the orientation result an anatomy. First, **rotation toward the human axis is uniformly tiny**: $\leq$ 3.5% of $\|\Delta s\|^2$ in every cell (stage means 1.1/2.3/2.8/1.8/1.9% for FS/FT/FT+PO/AT/CAL). The demonstration-anchored stages carrying the headline gain rotate no more than the weight-updating ones, so their advantage is not a subspace correction. Second, the structured part is **stretch** (17–30% on sarvam-m and gemma, tracking each stage's $\sigma$-ratio gain: added variance is mostly a louder base ranking). Aya isolates the mechanism by exception: its stretch is $\leq$ 0.1% and $\sim$97% of its update is off-axis reshuffling, which is why it moves $\sigma$ without converting it into Pearson. Across cells the residual is 67–99%: training *amplifies then reshuffles*, and the orientation axis is essentially never moved.

A complementary check isolates what the demonstrations contribute: *post-hoc* affine rescaling of the FT+PO judge onto human anchors never reproduces the gains (Fig. 11, Table 23), so the lift needs the demonstrations at inference, not a scale fix.

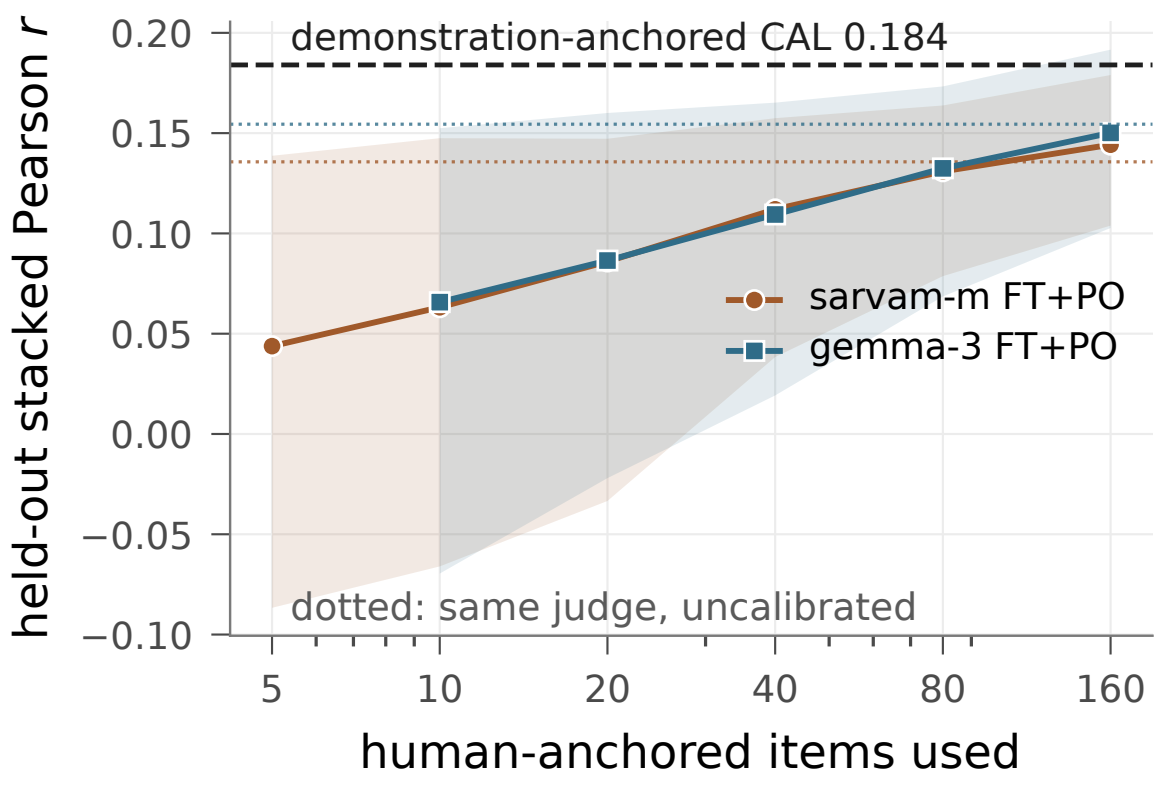

Figure 11: Held-out $r$ vs. anchor count; rescaling plateaus below CAL (0.184) and its interval spans zero under $\sim$40 anchors. Extended caption in Table 7.

| Construction | CH base | DLQ base |
|---|---|---|
| per-rubric, matched ref. *(body)* | 83.5–85.5 | 60.7/73.1/75.4 vs. floor 56.4/63.8/64.4 |
| per-rubric, 3-rater mean ref. | – | 58.5/71.6/74.0 |
| per-judge stacked | 88.3 | 84.3–88.9 vs. floor 73.9–78.1 |
| cohort stacked | 88.08 | 58.54 obj.; 80.17 subj. |
| cohort PCA axis → human | 79.0–85.9 | 47.7 (TQ); 64.3/65.5 (subj.) |

Table 4: **One cohort, four angle constructions** (TQ/HAL/LA where three values are given). Stacked angles take the *largest* principal angle, so they upper-bound the picture; the cohort PCA axis is the judges' best shared direction, so it lower-bounds it; per-rubric angles sit between. Human floors exist only on DLQ, which has three raters. Every body claim compares like with like inside one construction; the shaded row is the one the body uses. Full discussion in App. N.

## 6 Conclusion

High inter-judge agreement is the de facto license for replacing human evaluation with LLM judges. We make that license checkable, and it fails where it is most relied upon: on subjective rubrics our judges reach the human panel's *consensus* level while recovering only 58–66% of its *agreement*, collapsing onto a shared axis 64–66° from the human one; on the verifiable rubric the gap nearly closes, and training restores spread but not orientation. Because the check needs only human-anchored items and judge scores already in hand, it runs as a four-step protocol (App. B): test spread, rank, and angle against permutation nulls, reporting the three agreement legs together; hold each rater out to build the floor a judge must reach; trust an ensemble only if the shared axis itself passes; and on failure, anchor on human-rated demonstrations, not more judges. Consensus shows that judges share an axis; geometry shows whether it is the human one.

## Limitations

The geometric account is a diagnostic abstraction over observed score covariance, not a mechanistic claim about LLM internals (Sec. 3): we cannot tell whether judges fail to *perceive* human-weighted dimensions or perceive but down-weight them, and label noise, item ambiguity, and pretraining familiarity remain compatible explanations of the objective–subjective boundary, so our claim is predictive rather than causal. Consistent with this, $d_R$ separates the two regimes but is flat within each ($\rho = -0.03$ over six subjective cells) and is entangled with inter-rater agreement at $n = 9$ cells. The objective regime also rests on a single verifiable rubric, replicated across three domains and eight languages but still one rubric, and even there the cohort mean does not reach the human floor: the crossing we report is for the best single judge and for two-judge ensembles. Human references are uneven, since COMMUNITY-HEALTH merges 15 annotators into one pool without per-rater identifiers, so no inter-human floor or rater-level jackknife exists there and every matched-floor claim is scoped to the DLQ subset with three independent raters; CH angles additionally carry bootstrap intervals $6.3°$ wide against a $1.6°$ between-judge SD, so we rank no CH judges by angle. Claims are therefore scoped to the tested datasets, languages, and judge families: the direction of every effect held in all eight languages but its size varied by two orders of magnitude and effectively vanished on Telugu, per-language angles are underpowered at $n = 60$–$300$, training covers three open Indic-capable families (24–32B) on one benchmark with the demonstration-anchored result sarvam-m-only, and the diagnostics assume a fixed rubric basis. Finally, closed-API judges may change silently; we tabulate versions and call dates and release all raw outputs (App. O), but exact reproduction of API-judge scores cannot be guaranteed.

## Ethics Statement

Our considerations follow Bender and Friedman (2018). This study performs no new human-subjects data collection: we reuse two released benchmarks, COMMUNITY-HEALTH (Samiksha) (Hamna et al., 2026; Bhat et al., 2025) and REAL-WORLD CULTURAL QA (Pariksha) (Watts et al., 2024), whose construction and data collection protocols are documented in those papers. They contribute the benchmarks and their human labels; the present paper reuses that released data and contributes the geometric diagnostics and regime analysis built on it, with no overlap in stated contributions. The released data contain no sensitive personal information, and the items are advisory questions rather than patient records. The paper's central caution is itself a risk-mitigation contribution: deploying LLM judges without the geometric check of Sec. 6 risks silently optimising community-facing systems, such as health advice, toward a judge axis orthogonal to community needs. We used AI assistants as detailed in App. Q; all substantive analyses, evaluations, and interpretations were performed and verified manually by the authors.

## A Appendix guide

The appendix is organised so that each part answers one kind of question. Readers checking a specific claim can go directly to the relevant row of Table 5.

## B Extended discussion: the pre-trust protocol

Sec. 6 states the check in prose. This appendix gives the reading of the results that motivates it, and then the procedure itself.

**Subjective inter-LLM consensus is perspective collapse, not alignment.** On REAL-WORLD CULTURAL QA's subjective rubrics the cohort reaches the human panel's *consensus* level ($r_{\text{LL}}$ 0.356/0.349 vs. $r_{\text{HH}}$ 0.357/0.364) while agreeing with humans at only 58–66% of it (Table 1A). Raters legitimately differ on culturally embedded items; the LLMs match the consensus statistic not by recovering that perspective variance but by collapsing onto a shared axis 64–66° from the human one on DLQ (79–86° on CH), at effective rank $\leq 3$ where humans use the full basis (Sec. 5.3, 5.6). In Table 27 the cohort scores medical completeness while humans weight institutional vulnerability, affordability, and emotional reassurance. Reporting the three legs separately on one construction makes the collapse legible; an averaged human reference hides it.

**Training corrects spread, not orientation, and aggregate gains hide per-rubric harm.** Fine-tuning and FT+PO move $\sigma_{\mathcal{J}}/\sigma_H$ from 0.32 to 1.08 while the angle stays at 87–88°: rotate-$\rightarrow$-human is $\leq 3.5\%$ of $\|\Delta s\|^2$ in every stage cell, including the two demonstration-anchored stages that produce the largest Pearson gains (Sec. 5.8). Those stages also degrade completeness while over-spreading accuracy. On sarvam-m only, demonstration anchoring lifts all four rubrics, putting a calibrated 24B Indic judge (0.184) above gpt-5.5 (0.123) yet still far below DLQ inter-human agreement (0.474).

**The protocol, step by step.** The four-step check below is the procedure that follows from the readings above; the commentary after it explains why each step matters.

**Geometric check for LLM-judge consensus**

1. Compute $\sigma_{\mathcal{J}}/\sigma_H$, $r_{95}$, $\theta$ against a permutation null; report $r_{\text{LL}}$, $r_{\text{LH}}$, $r_{\text{HH}}$ together.
2. With $\geq 3$ raters, hold one out; score it *and* each judge against the rest. Rank-matched only if $\theta$ beats the null and nears that floor.
3. Ensembling stalls at the cohort's shared axis: it substitutes only if that axis passes.
4. On failure, anchor on human-rated demonstrations; rescaling does not help.

Step 2 is the one most often skipped, and it is the one that changes conclusions: scoring a judge against a multi-rater mean while scoring humans against each other flatters the judge by several degrees and can place it above a floor it does not actually reach (App. N). Budget for it, too: at half our item count the below-null verdict stopped replicating (App. M). Step 3 matters because a panel's headline number averages over judges that may share one axis; on our subjective rubrics an unlimited ensemble of the tested judges is bounded at 66.5°/64.6°, still outside the 64.4°/63.8° floor (Table 3). Step 4 is demonstration anchoring rather than rescaling: purely post-hoc affine maps never reach the calibrated judge, and below $\sim 40$ anchors their interval spans zero (Fig. 11).

## C Extended figure captions

Main-body figures carry short captions to keep the analysis compact. The full reading of each, including construction, references, and the caveats attached to it, is given here.

## D The geometric lens, expanded

This section expands Sec. 3. It adds the conceptual picture that motivates the diagnostics, works the variance identity through to the measured magnitudes, and states what the lens does and does not claim.

**An idealised picture (conceptual, not mechanistic).** Imagine quality as a vector with stylistic, factual, and culturally situated components, weighted rubric-conditionally by a rater: objective rubrics concentrate weight on the factual component, subjective rubrics spread it onto culturally situated ones (Hershcovich et al., 2022). A judge whose scores vary along few directions behaves, *at the level of its score matrix*, like a low-rank map. The picture organises the predictions P1–P6 and nothing more: we do not model the judge as a literal projection operator, and no measurement in this paper depends on that reading. Where the text says a cohort "shares an axis," the evidence is a principal-component decomposition of observed scores (Sec. 5.6), not a claim about model internals.

| App. | Contents | Answers the question |
|---|---|---|
| A | This guide | What is where in the appendix? |
| B | The pre-trust protocol, in full | How do I actually run the geometric check on my own judge and rubric? |
| C | Extended figure and table captions | What exactly do the main-body figures and tables show? |
| D | The geometric lens, expanded | What exactly does Sec. 3 claim, and what does it not? |
| E | Per-domain breakdown: datasets, languages, generators, judges | How many items, languages, and judges are in each cell, and how do the totals add up? |
| F | Judge roster, trained-variant roster | Which models judge which benchmark, and what is every trained variant? |
| G | Rubric definitions and ordinal scales | What exactly was each rubric asking, and on what scale? |
| H | Results-grid column key and codes | How do I read Table 1? |
| I | Verbatim evaluation prompts | What were the judges actually shown? |
| J | Variant recipes and hyper-parameters | How was each trained judge produced? |
| K | Boundary condition and scope | Which rubrics fall in which regime, and what does the lens *not* claim? |
| L | Angle methodology, permutation baseline | How is every angle computed, and against what null? |
| M | Robustness and control tables | Do the results survive ordinal, reference, FDR, self-preference, and split-half controls? |
| N | Reconciling the four angle constructions | Why do angles for the same cohort differ by 30° across tables? |
| O | Reproducibility, artifact index, API versions | How do I regenerate every number in this paper? |
| P | Qualitative item study | What do the two regimes look like at the level of single items? |
| Q | Use of AI assistants | Where were LLM-based tools used, and how was their output checked? |

Table 5: **Reader's guide to the appendix.**

**From the identity to the measured magnitudes.** Equation 2 is an identity, not an analogy: for a single direction $\cos\theta = |r|$, so $1 - \cos^2\theta = 1 - r^2$ is exactly the fraction of human score variance that the judge direction leaves unexplained. This wires the lens to quantities we can measure. On task quality one judge direction leaves $1 - 0.418^2 = 83\%$ of human variance unexplained; the entire judge cohort, given a cross-validated ridge fit, still leaves 55% (App. M). Aggregation therefore recovers part of the residual but not most of it, which is the same conclusion the ensemble analysis reaches geometrically (Sec. 5.6). On the subjective DLQ rubrics the cohort leaves $\approx 80\%$ unexplained, and on COMMUNITY-HEALTH essentially all of it (CV $R^2 \leq 0$).

**Why rank is necessary but not sufficient.** A cohort whose scores span $k_{\mathcal{J}}$ directions cannot reproduce human variation that needs $d_R$ of them when $k_{\mathcal{J}} < d_R$, which makes $k_{\mathcal{J}} \geq d_R$ necessary. It is not sufficient because rank is a budget, not an aim: a judge may spend four directions on distinctions humans do not make. Our data separate the two cleanly. Across the 40 CH judges effective rank is uncorrelated with alignment ($r = 0.12$; $r_{95}$ spans 1–4 while angles span 85.0–89.9°), whereas the objective rubric stays compressed ($\sigma$-ratio 0.687) yet rotates to within 4.3° of the matched human floor. Orientation, not capacity, is what tracks agreement, which is why we state the boundary as an overlap condition and treat blindness risk as increasing with $1 -$ overlap rather than with a rank deficit alone.

## E Per-domain breakdown of datasets, languages, generators, and judges

Per-domain coverage (referenced from Sec. 4). Table 8 lists languages, item counts, and rubrics per domain; Table 9 lists answer generators and judge variants per domain. Trained judge variants apply only to COMMUNITY-HEALTH. Every count in these tables is regenerated from the released CSVs by one script (App. O).

**Human reference panels.** COMMUNITY-HEALTH: the released labels carry no per-rater identifiers and have only sparse per-rater overlap, so ratings are merged into one HUMAN-COMMUNITY-HEALTH refer-ence (which is why no CH inter-human floor exists; see Limitations). REAL-WORLD CULTURAL QA: three independent raters ($h_1$, $h_2$, $h_3$) per item per (domain, language), giving a measurable H–H reference.

| Figure | Extended caption |
|---|---|
| Fig. 2 *Agreement-gap map* | One row per rubric×dataset cell: red = mean LLM–human $r$, blue = mean inter-LLM $r$, black ticks = inter-human $r$ (DLQ only); segment length = the LL−LH gap. All three legs are pairwise, so the three markers are directly comparable. Subjective CH rubrics gap widest; task quality narrows it most. |
| Fig. 4 *Per-judge signature (CH)* | All 40 CH judges in ($\sigma$-ratio, angle-to-human) space, against the single merged human reference (no inter-human floor exists on CH, so no floor line is drawn). Colour and marker give the judge family (same mapping as Figs. 12, 6), size scales with $r_{95}$, grey dots are trained variants. Per-judge angles here are not separable from one another (Sec. 5.4); the figure shows the cohort's position and spread, not a ranking. |
| Fig. 5 *Rubric-conditional geometry (DLQ)* | One panel per rubric in ($\sigma$-ratio, angle) space, both against the matched two-rater reference: dots = the five base judges, × = the three held-out humans, + = the five-judge ensemble mean, shaded = judge density. This is the domain-pooled view of Fig. 6's columns: task quality pulls the judges toward the human band while $\sigma$ stays below 1, and the subjective panels leave a clear separation. |
| Fig. 12 *Per-language signature (DLQ)* | $n$ = 60–300 per panel; total 1,182. Each panel plots the five base judges in ($\sigma_{\mathcal{J}}/\sigma_H$, angle) space, both measured against the matched two-rater reference: colour = judge category, cream band = the $\sigma < 1$ compression zone, dashed green = the held-out-rater human floor, open diamond = the five-judge ensemble mean. Compression is universal, but the angle axis is underpowered at these cell sizes and saturates near 90° for judges and humans alike, so we draw no per-language angle conclusions (Sec. 5.5); note also that the diamond averages five judges while the floor is a single rater, so it is favoured by construction. |
| Fig. 7 *Inter-judge subspace similarity (CH)* | Pairwise cosines between Z-scored item×rubric vectors of the 14 base judges + human, Ward-reordered; side strips = family; the human row is outlined and uniformly pale. $\overline{\cos}_{LL} = 0.16$ vs. $\overline{\cos}_{LH} = 0.08$; the deepest cells are qwen3-next↔gpt-4o (0.72), llama-3.3↔mistral-L2 (0.67), llama-4-scout↔mistral-L3 (0.54), then the qwen3 pair (0.50) — the first three cross-vendor, which is why Sec. 5.6 reads the shared orientation as cohort-wide rather than familial. |

Table 6: **Extended captions for the main-body figures** (part 1).

## F Judge models

Table 10 lists the 15 base LLM judges; Table 11 enumerates the 27 trained variants over three Indic-capable families, plus the 4 human reference pools used as anchors. Four judges (deepseek-r1, grok-4, mistral-large-3, llama-3.3-70b) act as cross-dataset anchors.

## G Rubrics and scoring scales

Table 12 gives per-dataset rubric definitions and ordinal scales (Sec. 4). The COMMUNITY-HEALTH comparative paradigm returns one of $\{A, B, \text{NOT SURE}\}$ per item with a one-sentence English rationale. Verbatim prompts in App. I.

## H Additional figures

### H.1 Consolidated results grid: column key and codes

Table 13 is a compact legend for Table 1. Per-panel narrative summary in Table 14.

## I Evaluation prompts

Two prompts: STANDALONE multi-rubric and COMPARATIVE pairwise. Both are language-agnostic; the judge evaluates within the answer's language and writes the explanation in English.

**STANDALONE MULTI-RUBRIC SCORING PROMPT**

**Task.** Evaluate the answer to the question on four rubrics.
**Audience.** Internal evaluator. Do not browse. Use only general knowledge and the given text.
**Language.** The (question, answer) pair may be in Hindi, Kannada, Malayalam, Bengali, Tamil, Telugu, Marathi, or Odia. Judge *within* the answer's language. Write the explanation in concise English.

**Rubrics** (one rating per rubric: `Yes` / `Somewhat` / `No`):

*i.* **Clarity & fluency**: is the language clear, grammatically correct, easy to understand?

*ii.* **Helpfulness & relevance**: does the answer directly

| Float | Extended caption |
|---|---|
| Fig. 8 *Stage-wise bootstrap* | $B = 1000$ over the 405 CH core items: rows are the three diagnostics, columns the three families; violins are bootstrap distributions, points stage means. Stages are grouped by intervention type — the prompt- and demonstration-based ones first (base, FS, AT, CAL), then the weight-updating ones (FT, FT+PO) — which is neither pipeline order nor the order of their Pearson gains; Fig. 10 instead uses pipeline order. Spread rises reliably; angle distributions overlap throughout. |
| Fig. 9 *Per-rubric training progression* | Best in-stage judge per family; one line per COMMUNITY-HEALTH rubric. *Top:* Pearson $r$, dotted line = the gpt-5.5 anchor. *Bottom:* $\sigma$-ratio, dotted line = judge spread equal to human spread. Fine-tuning lifts accuracy while pushing its spread past $2\times$ the human range and dragging completeness down; only the demonstration-anchored stages (AT, CAL) raise every rubric with spread still inside the human range. |
| Fig. 10 *Training-delta decomposition* | Per-(family, stage) decomposition of $\|\Delta s\|^2$ into stretch along the base axis (blue), rotation toward the human axis after the base direction is removed (green), and off-axis residual (grey); rubric-mean over the four CH rubrics. Stages run base$\to$FS$\to$FT$\to$FT+PO$\to$AT$\to$CAL, the pipeline order, which is not the order of their Pearson gains. Rotation toward human is $\leq 3.5\%$ across all eleven stage cells, including both demonstration-anchored stages; aya's near-zero stretch with $\sim$97% residual is the clearest case of variance added without alignment. |
| Fig. 11 *Anchor-size curve* | Held-out stacked Pearson $r$ after fitting per-rubric affine maps of two FT+PO judges on $n$ human-anchored items (500 resamples; bands are 95% intervals). Dotted lines: the same judges uncalibrated. Dashed line: the demonstration-anchored calibrated judge (0.184). Rescaling improves monotonically with anchors but plateaus below the uncalibrated judge on gemma-3 and below CAL on both, and under $\sim$40 anchors its interval spans zero: the calibration gain of Sec. 5.7 is not a scale correction that more anchors would buy. |
| Table 23 *Rescaling by anchor count* | Held-out stacked Pearson $r$ after fitting per-rubric affine maps of the FT+PO judge onto $n$ human-anchored items (500 resamples, seed 20260828; "–" = degenerate fits at $n$=5). Rescaling never reaches the demonstration-anchored calibrated judge (0.184) and *hurts* below $\sim$40–80 anchors; on gemma-3 it never beats the uncalibrated judge at all. The calibration lift therefore requires the human-rated demonstrations at inference, not a scale correction. |
| Table 3 *Ensemble-size curve* | Mean angle° over all $\binom{5}{k}$ subsets of the five base judges, each scored against the same held-out two-rater mean that defines the floor in the last column. $k\to\infty$ sums the geometric tail of the decrements, which shrink by a near-constant factor (0.59–0.62). On task quality two judges already pass the floor and the limit is $6^\circ$ inside it; on both subjective rubrics no attainable ensemble reaches the floor, and the extrapolated limit stays outside it. Note the asymmetry that runs against us: the ensemble averages five judges while the human side is a single rater. |

Table 7: **Extended captions for the main-body figures and tables** (part 2).

address the question and provide useful information?

*iii.* **Accuracy (general perception)**: based on general knowledge, does the information seem trustworthy and factually correct?

*iv.* **Completeness & conciseness**: does the answer provide sufficient detail without being verbose or repetitive?

**Scoring.** `Yes`= 2, `Somewhat`= 1, `No`= 0. Overall = sum of the four.

**Output schema (strict, single minified JSON, no prose, no code fences;** line breaks below are for display only**):**

```
{
 "clarity": {"rating":"Yes|Somewhat|No",
           "explanation":"..."},
 "helpfulness": {"rating":"Yes|Somewhat|No",
               "explanation":"..."},
 "accuracy": {"rating":"Yes|Somewhat|No",
            "explanation":"..."},
 "completeness": {"rating":"Yes|Somewhat|No",
                "explanation":"..."},
 "scores": {"clarity":int,"helpfulness":int,
               "accuracy":int,"completeness":int,
               "overall":int}
}
```

**Question:** `{{question}}`
**Answer:** `{{answer}}`

Return the JSON now.

On REAL-WORLD CULTURAL QA the same template substitutes three rubrics: **task quality**, **linguistic acceptability** ($\{0, 1, 2\}$), **hallucinations** ($\{0, 1\}$).

### COMPARATIVE (PAIRWISE PREFERENCE) PROMPT

**Task.** Compare two answers to the same question and pick the better one overall.
**Audience.** Internal evaluator. Use only the question, the two answers, and general world knowledge; no external browsing.
**Languages.** The (question, A, B) triple may be in Hindi, Kannada, or Malayalam. Judge within the answers' lan-

| Domain dataset | Languages | Qs | QA | Rubrics (eval type) |
|---|---|---|---|---|
| COMMUNITY-HEALTH (Hamna et al., 2026; Bhat et al., 2025) | Hi; Kn; Ml | 810 | 2,430 | clarity, helpfulness, accuracy, completeness (STANDALONE, $\{0, 1, 2\}$); A/B/NS (COMPARATIVE, 2,430 pairs) |
| DLQ: HEALTH & WELLNESS (Watts et al., 2024) | 8 Indic† | – | 284 | task quality ($\{0, 1, 2\}$, *obj.*); linguistic acceptability ($\{0, 1, 2\}$); hallucinations ($\{0, 1\}$) (STANDALONE) |
| DLQ: FINANCE (Watts et al., 2024) | 8 Indic† | – | 323 | same as above |
| DLQ: EVERYDAY LIFE (Watts et al., 2024) | 8 Indic† | – | 575 | same as above |
| **Totals** | **8 unique langs** | **968** | **3,612** | **4 domain datasets; 170,890 LLM-judge calls** |

Table 8: **Per-domain breakdown (datasets, languages, items, rubrics).** *Hi/Kn/Ml* = Hindi/Kannada/Malayalam. †DLQ languages: Bengali, Hindi, Kannada, Malayalam, Marathi, Odia, Tamil, Telugu; per-language item counts are uneven (Hindi 300, Telugu 166, Tamil 150, Bengali 144, Malayalam 130, Kannada 121, Marathi 111, Odia 60). DLQ retains the 1,182 items with three complete human ratings, drawn from 158 questions across 27 answer generators (per-question counts vary, so no per-domain question total is defined). COMMUNITY-HEALTH: 270 questions $\times$ 3 languages $\times$ 3 answer generators = 2,430 QA pairs; all geometric/training analyses use its 405-item core (45 questions/language $\times$ 3 languages $\times$ 3 generators) with complete topic metadata (Sec. 4). DLQ's separately collected comparative split (21,610 pairs over 10 languages, adding Gujarati and Punjabi) is excluded from analysis (App. M.12).

| Field | COMMUNITY-HEALTH |
|---|---|
| Answer generators | 3: llama-3.1-405b, qwen3, sarvam-m (Hamna et al., 2026; Bhat et al., 2025) |
| Judges, *base* (14) | deepseek-r1; gpt-5.5; gpt-4o; grok-4; mistral-large-3; mistral-large-2; llama-4-scout-17b; llama-3.3-70b; qwen3-235b; qwen3-next-80b; krutrim-2-instruct; aya-expanse-32b; gemma-3-27b; sarvam-m-24b |
| Judges, *improved sarvam-m* (12 standalone + 1) | few-shot; fine-tune v1/v2/v3; pref-optim v2 (v1 comparative-only); fine-tune+pref-optim v1/v2; demo-SFT v1/v2; demo-SFT v1-inf3; calibrated v2/v3 |
| Judges, *improved gemma-3* (7) | few-shot; fine-tune v1/v1-b128/v2; pref-optim v1/v2; fine-tune+pref-optim v2 |
| Judges, *improved aya-expanse* (7) | few-shot; fine-tune v1/v2; pref-optim v1/v2; fine-tune+pref-optim v1/v2 |
| **Field** | **DLQ: HEALTH & WELLNESS / FINANCE / EVERYDAY LIFE** (identical roster across all three) |
| Answer generators | 27 Indic and general instruction-tuned models (Watts et al., 2024) |
| Judges, *base* (5) | deepseek-r1; gpt-4; grok-4; mistral-large-3; llama-3.3-70b |
| Judges, *improved* | none (trained variants not applied to DLQ) |

Table 9: **Answer generators and judge variants per domain.** COMMUNITY-HEALTH totals 14 base + 27 trained variants (26 in the standalone paradigm; sarvam-m pref-optim v1 appears only in the comparative paradigm); the three DLQ domains share the same 5 base judges, of which four (deepseek-r1, grok-4, mistral-large-3, llama-3.3-70b) also judge COMMUNITY-HEALTH. Combined: **42 distinct LLM judges** (15 base + 27 trained) across the study.

guage; write the explanation in concise English ($\leq$ 40 words).

**Holistic criteria.** Clarity & fluency, helpfulness & relevance, factual accuracy / plausibility, completeness (no critical omissions), and safety. Prefer the answer that better serves the user overall; do not reward hallucinated or unsafe content.

**Decision rule.** Choose `A` or `B` if one answer is even slightly better overall. Choose `Not sure` only if both are essentially equal or both are unintelligible/empty.

**Output (strict, single JSON, exact keys):**

```
{
  "winner": "A" | "B" | "Not sure",
  "explanation":
    "<concise justification <=40 words>"
}
```

**Question:** `{{question}}`
**Answer A:** `{{answer_a}}`
**Answer B:** `{{answer_b}}`

Return only the JSON now.

| Model | Family / orientation | Size | Access | Data |
|---|---|---|---|---|
| gpt-5.5 | GPT (OpenAI), general | n/d | API | CH |
| gpt-4o | GPT (OpenAI), general (OpenAI, 2024) | n/d | API | CH |
| gpt-4 | GPT (OpenAI), general (OpenAI et al., 2023) | n/d | API | DLQ |
| grok-4 | Grok (xAI), general | n/d | API | both |
| deepseek-r1 | DeepSeek (DeepSeek-AI, 2025), reasoning (open weights) | 671B | API | both |
| mistral-large-3 | Mistral, general | 123B | API | both |
| mistral-large-2 | Mistral, general | 123B | local | CH |
| qwen3-235b | Qwen (Qwen Team, 2025), multilingual | 235B | local | CH |
| qwen3-next-80b | Qwen (Qwen Team, 2025), multilingual | 80B | local | CH |
| llama-3.3-70b | Llama (Grattafiori et al., 2024), general | 70B | both | both |
| llama-4-scout-17b | Llama, general (17B active) | 109B | local | CH |
| aya-expanse-32b | Aya (Cohere) (Dang et al., 2024), multilingual | 32B | local | CH |
| gemma-3-27b | Gemma (Gemma Team, 2025), multilingual | 27B | local | CH |
| sarvam-m-24b | Sarvam (Sarvam AI, 2025), Indic-focused | 24B | local | CH |
| krutrim-2-instruct | Krutrim, Indic-focused | 12B | local | CH |

Table 10: The 15 base LLM judges. Size = reported total parameters where disclosed (n/d = not disclosed). Access/Data "both" on the llama-3.3-70b row is not a repeated cell: that judge was served over the API for DLQ and locally for CH, and it scores both benchmarks. llama-3.1-405b and qwen3 appear in the study only as CH *answer generators*, never as judges. Per-domain coverage in Table 8; trained variants in App. J; API deployments and call dates in App. O.

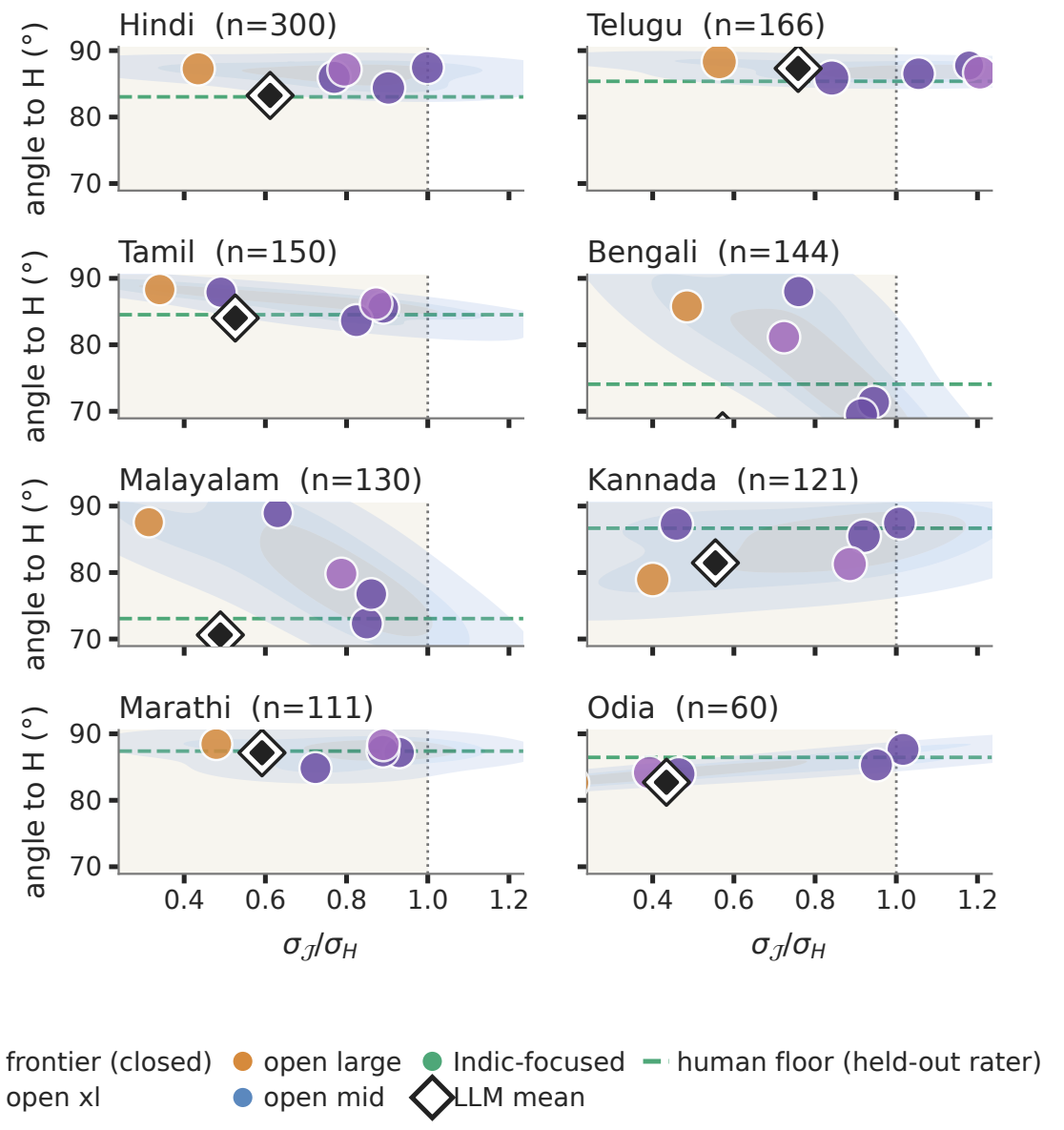

Figure 12: **Per-language signature (DLQ).** Compression is universal; angles are underpowered here, so we read none. Each panel plots the five base judges in $(\sigma_{\mathcal{J}}/\sigma_H$, angle) space against the matched two-rater reference; the dashed line is the held-out-rater floor and the open diamond the five-judge ensemble. At $n = 60$–$300$ items per language the stacked angle saturates near 90° for judges and humans alike, which is why Sec. 5.5 draws no per-language angle conclusion and reads the agreement gap instead.

# J Judge variant recipes

Six recipes applied across three open base families on COMMUNITY-HEALTH. Per-family roster in Table 11. Common controls for all judge scoring: greedy decoding (temperature 0.0, top-$p$ 1.0), max 512 output tokens (1024–2048 on parse-retry passes), vLLM 0.18.0 in bf16 on 8×B200; training seed 42; bootstrap seed 20260525. SFT labels are 5,683 per-rubric human judgements from the CH source study (train/val 5,114/569, stratified by language×rubric×rating; v2 runs oversample rating classes to balance), and DPO uses 2,935 strict-preference comparative triples (NOT SURE dropped; 90/10 split); the training questions have *zero overlap* with the 810 evaluation questions. Per-version YAML configs and training logs ship in the release (App. O).

# K Boundary condition and scope of the framework

Extends Sec. 3. Let $d_R$ be the participation ratio of the human score matrix under rubric $R$ (Eq. 1, measured from human raters alone and capped at the panel size, so it reports the dimensionality of rater *disagreement* rather than of human quality itself) and $k_{\mathcal{J}}$ the judge cohort's effective rank. The boundary is *rank-necessary, orientation-decisive* (Eq. 2): $k_{\mathcal{J}} \geq d_R$ is required for alignment to be possible, but the decisive quantity is the orientation overlap $\cos^2\theta(\mathcal{S}_{\mathcal{J}}, \mathcal{S}_H)$; measured effective rank does not predict alignment ($r = 0.12$ across 40 CH judges; Sec. 5.4). Table 16 summarises which regimes the framework does and does not cover.

| Variant ID | Recipe | Display | Stacked $r$ |
|---|---|---|---|
| `sarvamm` | base | sarvam-m (24b) | 0.106 |
| `sarvamm_fs` | few-shot | sarvam-m (fewshot) | 0.184 |
| `sarvamm_ft_v1` | fine-tune | sarvam-m (FT v1) | 0.119 |
| `sarvamm_ft_v2` | fine-tune | sarvam-m (FT v2) | 0.127 |
| `sarvamm_ft_v3` | fine-tune | sarvam-m (FT v3) | 0.120 |
| `sarvamm_po_v1` | pref-optim | sarvam-m (PO v1; comparative-only) | — |
| `sarvamm_po_v2` | pref-optim | sarvam-m (PO v2) | 0.094 |
| `sarvamm_ft_po_v1` | FT + PO | sarvam-m (FT+PO v1) | 0.112 |
| `sarvamm_ft_po_v2` | FT + PO | sarvam-m (FT+PO v2) | 0.136 |
| `sarvamm_ft_v3_demo1` | demo-SFT | sarvam-m (demo-SFT v1; AT v1) | 0.106 |
| `sarvamm_ft_v3_demo2` | demo-SFT | sarvam-m (demo-SFT v2; AT v2) | 0.160 |
| `sarvamm_ft_v3_demo1_infer3` | demo-SFT | sarvam-m (demo-SFT v1-inf3; AT v1-inf3) | 0.177 |
| `sarvamm_po_v3_cal_2demo` | calibrated | sarvam-m (CAL v2) | 0.178 |
| `sarvamm_po_v3_cal_3demo` | calibrated | sarvam-m (CAL v3) | 0.184 |
| `gemma-3-27b-it` | base | gemma-3 (27b) | 0.124 |
| `gemma-3-27b-it_fs` | few-shot | gemma-3 (fewshot) | 0.166 |
| `gemma-3-27b-it_ft_v1` | fine-tune | gemma-3 (FT v1) | 0.102 |
| `gemma-3-27b-it_ft_v1_b128` | fine-tune | gemma-3 (FT v1-b128) | 0.090 |
| `gemma-3-27b-it_ft_v2` | fine-tune | gemma-3 (FT v2) | 0.164 |
| `gemma-3-27b-it_po_v1` | pref-optim | gemma-3 (PO v1) | 0.115 |
| `gemma-3-27b-it_po_v2` | pref-optim | gemma-3 (PO v2) | 0.115 |
| `gemma-3-27b-it_ft_po_v2` | FT + PO | gemma-3 (FT+PO v2) | 0.155 |
| `aya-expanse-32b` | base | aya-expanse (32b) | 0.011 |
| `aya-expanse-32b_fs` | few-shot | aya-expanse (fewshot) | 0.019 |
| `aya-expanse-32b_ft_v1` | fine-tune | aya-expanse (FT v1) | 0.041 |
| `aya-expanse-32b_ft_v2` | fine-tune | aya-expanse (FT v2) | 0.098 |
| `aya-expanse-32b_po_v1` | pref-optim | aya-expanse (PO v1) | 0.072 |
| `aya-expanse-32b_po_v2` | pref-optim | aya-expanse (PO v2) | 0.072 |
| `aya-expanse-32b_ft_po_v1` | FT + PO | aya-expanse (FT+PO v1) | 0.041 |
| `aya-expanse-32b_ft_po_v2` | FT + PO | aya-expanse (FT+PO v2) | 0.107 |
| `h_ch` | human reference | CH human reference (merged pool) | ref. |
| `h1` | human rater | DLQ human $h_1$ | — |
| `h2` | human rater | DLQ human $h_2$ | — |
| `h3` | human rater | DLQ human $h_3$ | — |

Table 11: The three trainable families and their 27 trained variants ($12 + 7 + 7 = 26$ in the standalone paradigm; `sarvamm_po_v1` appears only in the comparative paradigm), plus 4 human reference pools. Together with the 15 base judges of Table 10 the study spans 42 distinct LLM judges. **Stacked** $r$ is each variant's stacked Pearson correlation with the merged CH human pool over all four rubrics ($n = 1620$ judge–item scores), the same quantity plotted in Fig. 13; higher is better, and even the top value is far below DLQ inter-human agreement (0.474). "ref." marks the pool the correlation is taken against; "—" marks rows with no standalone CH score (the comparative-only variant and the three DLQ annotators). Two cautions when reading this column. First, it is an *aggregate*: few-shot (0.1837) and CAL v3 (0.1845) are indistinguishable here, and the demonstration-anchored advantage is visible only per rubric, where few-shot leaves completeness flat (Fig. 9). Second, identical entries within a family (e.g. gemma PO v1/v2) are genuine rather than duplicated cells: those variants are so compressed that they place nearly every item in the top rating class. Recipe definitions and hyper-parameters in App. J.

## L Angle methodology and permutation baseline

**Matrices.** For judge $\mathcal{J}$, the score matrix $\mathbf{S}_{\mathcal{J}} \in \mathbb{R}^{n\times m}$ holds ordinal scores for $n$ items and $m$ rubrics ($m$=4 on COMMUNITY-HEALTH over the 405-item core; $m$=3 on REAL-WORLD CULTURAL QA over 1,182 items). The human reference is the merged pool on CH (a single reference, so no floor is measurable there). On DLQ every comparison that is read against a human floor uses the *matched* reference: one of the three raters is held out, the mean of the other two is the reference, and the held-out rater and each judge are both scored against it, averaged over the three folds. This keeps both sides a single rater facing the same two-rater target; scoring judges against the three-rater mean while scoring humans against each other gives judges a less noisy target and shifts them several degrees in their own favour (App. N). Where an unmatched construction is reported for continuity with earlier tables, it is labelled as such.

**Per-rubric angle.** For one rubric, judge and reference are vectors $\mathbf{s}, \mathbf{h} \in \mathbb{R}^n$. After mean-centering, $\cos\theta = |\langle \mathbf{s}_c, \mathbf{h}_c\rangle| / (\|\mathbf{s}_c\|\|\mathbf{h}_c\|) = |r(\mathbf{s}, \mathbf{h})|$, so the per-rubric angle is exactly $\theta = \arccos|r|$ for Pearson $r$: per-rubric angles restate correlations in a geometry that composes with the stacked construction.

| Dataset | Rubric | Scale | Definition |
|---|---|---|---|
| COMMUNITY-HEALTH | clarity & fluency | $\{0,1,2\}$ | language clear, grammatical, easy to understand |
| COMMUNITY-HEALTH | helpfulness & relevance | $\{0,1,2\}$ | directly addresses the question with useful info |
| COMMUNITY-HEALTH | accuracy | $\{0,1,2\}$ | trustworthy under general medical knowledge |
| COMMUNITY-HEALTH | completeness & concis. | $\{0,1,2\}$ | sufficient detail without verbosity (short name: *completeness*) |
| REAL-WORLD CULTURAL QA | task quality (*obj.*) | $\{0,1,2\}$ | satisfies task constraints; verifiable ground truth |
| REAL-WORLD CULTURAL QA | linguistic acceptability | $\{0,1,2\}$ | native-speaker grammaticality and naturalness |
| REAL-WORLD CULTURAL QA | hallucinations | $\{0,1\}$ | binary presence of unsupported content |

Table 12: Rubric definitions and ordinal scales. `Yes`$=2$ / `Somewhat`$=1$ / `No`$=0$ on COMMUNITY-HEALTH; task quality is the single objective rubric in this study.

| Symbol / code | Meaning |
|---|---|
| $n$ | judges/humans in cohort |
| $\sigma_{\mathcal{J}}/\sigma_H$ | score-spread ratio (judge SD over human-reference SD, per rubric, item-level) |
| $\theta^\circ$ | angle to human reference (panel A: per-rubric arccos $\lvert r \rvert$ against the matched two-rater reference; panels B–C: per-judge stacked; App. L) |
| $\theta^\circ_{\text{HH}}$ / $r_{\text{HH}}$ | human floor: held-out rater vs. the same two-rater reference / pairwise inter-human Pearson (three DLQ raters; CH has one merged pool, so neither exists) |
| $r$ / $r_{\text{ref}}$ | Pearson vs. the panel's stated reference; in panel A all three legs are pairwise |
| $\Delta$ | gap to the panel reference |
| $r^{\mathcal{J}}_{95}$ | effective rank (95% variance) of the item$\times$rubric matrix |
| 4 | lifts all four CH rubrics over BASE (unmarked trained rows do not) |
| **CH / DLQ** | COMMUNITY-HEALTH (Samiksha) / REAL-WORLD CULTURAL QA (Pariksha) |
| **LA / HAL / TQ** | linguistic acceptability, hallucinations (subj.); task quality (obj.) |
| **FE / OXL / OM / IF** | CH base strata: **FE** API frontier = {gpt-5.5, gpt-4o, grok-4, deepseek-r1}; **OXL** open $\geq$100B = {mistral-large-3, mistral-large-2, qwen3-235b, llama-4-scout}; **OM** open 27–80B = {llama-3.3-70b, qwen3-next-80b, gemma-3-27b, aya-expanse-32b}; **IF** Indic-focused = {sarvam-m-24b, krutrim-2}; figure legends use a finer five-way scheme that splits OM into open-large (70–80B) and open-mid (27–32B) |
| **BASE / FS / FT / FT+PO / AT / CAL** | base; few-shot; LoRA SFT; SFT+DPO; demonstration-augmented SFT; demonstration-anchored calibration (AT, CAL: sarvam-m only) |

Table 13: Column key and code legend for the consolidated results grid.

| Panel | One-line narrative |
|---|---|
| (A) | the LLM–human shortfall below inter-LLM agreement is +0.087 on CH's compressed scale and +0.110–+0.149 across the DLQ rubrics; DLQ task quality closes most of the gap, reaching 88% of the inter-human band ($r = 0.418$ vs. 0.474) at a matched angle of 60.7° against a 56.4° floor, with the best single judge inside it. |
| (B) | no base stratum closes more than 0.04 of the CH gap; angles stay at 87.5–89.6° across all strata: base-model choice is not the lever. |
| (C) | $\sigma_{\mathcal{J}}/\sigma_H$ rises $0.319 \to 1.078$ across FT/FT+PO; $\theta^\circ$ stays at 87–88°; only the demonstration-anchored stages (AT, CAL) lift all four rubrics. |

Table 14: Per-panel narrative of Table 1.

**Stacked angle.** Columns of $\mathbf{S}_{\mathcal{J}}$ and $\mathbf{S}_H$ are mean-centered; each matrix is reduced by SVD to the smallest basis capturing 95% of squared singular-value energy; the reported statistic is the *largest* principal angle (Björck and Golub, 1973; Knyazev and Argentati, 2002) between the two bases, truncated to the smaller of the two ranks. Sensitivity: at 90/95/99% energy the CH cohort angle is 88.0/88.1/88.6°; raw, centered, z-scored, and rank-normalised preprocessing give 88.08/88.08/88.50/88.38°; the DLQ subjective and objective stacked angles (80.17°, 58.54°) are unchanged to two decimals under raw/centered/z-scored preprocessing at every cutoff (rank-normalised: 80.42°, 59.65°).

**Permutation null.** Because near-orthogonality is generic in high dimension, every angle is read against a permutation null: the human vector (per-rubric; 500 permutations per judge, seed 20260710) or the rows of the human matrix (stacked; 300 permutations) are shuffled and the angle recomputed. We report the null mean and the observed angle's percentile within the null. On DLQ, all 15 judge×rubric observed angles fall below *all* 500 shuffles, i.e., $p < 1/501 < 0.002$ for each test separately (nulls 88.6–88.7°); the 15 tests share the same item set, so they are not independent, but the weakest of them already survives a Bonferroni correction across all 15 ($0.002 \times 15 < 0.05$); the same holds for all 5 judges on the stacked objective and subjective constructions. On CH the cohort mean (88.08°) sits near the stacked null (88.80°), the strongest collapse in the study. Because Benjamini–Hochberg over 40 judges needs $p \le .00125$ at rank 1, below what 300 shuffles can resolve, we rerun the CH judge-level test at 20,000 shuffles: 11 judges are below the null at $p < .05$ against 2 expected by chance, 6 survive at $q < .05$ and 10 at $q < .10$ (Table 19). The ten base judges not below the null include every Indic-focused and open-mid judge.

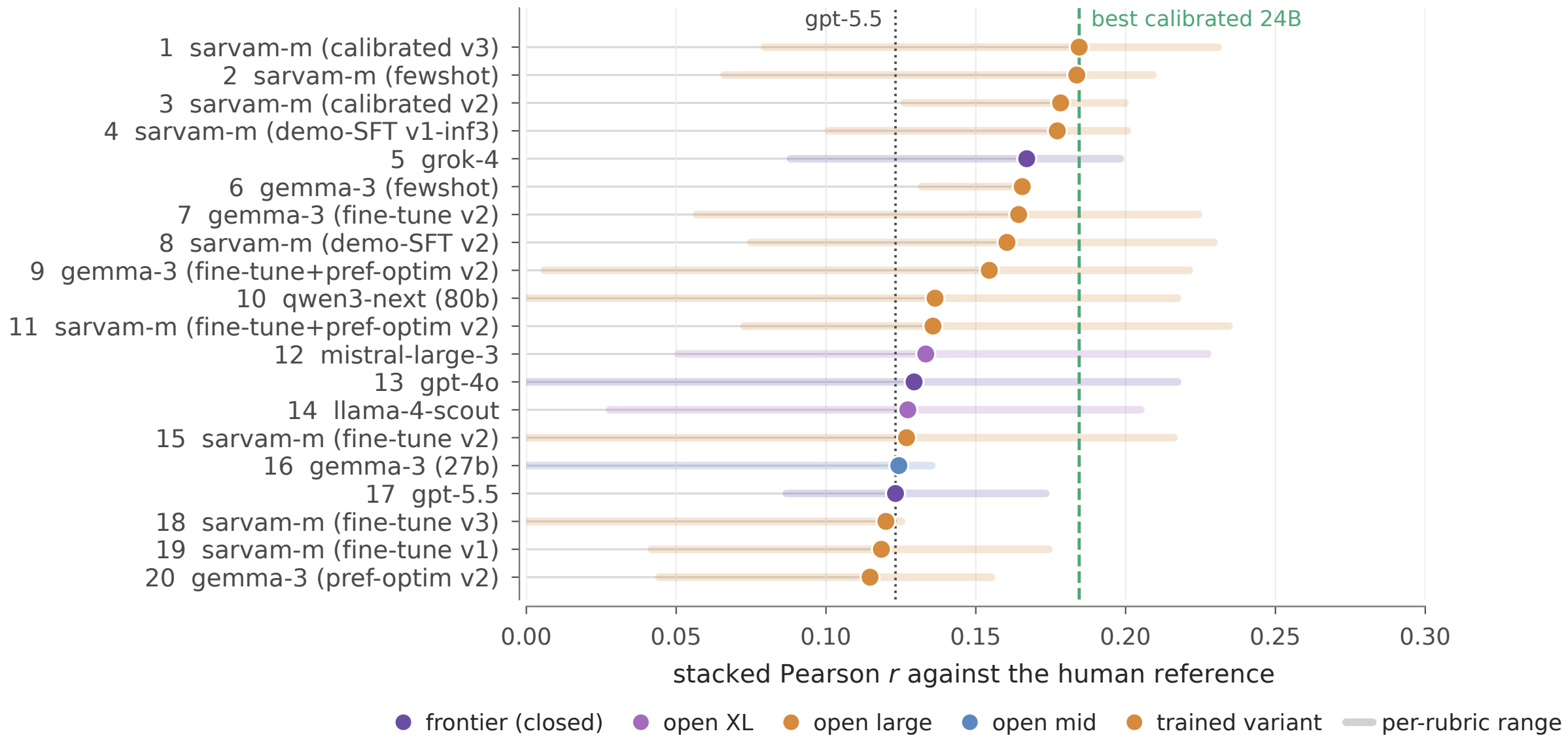

Figure 13: **Judge ranking.** Top 20 of the 40 COMMUNITY-HEALTH judges by stacked Pearson $r$ over the 405-item core ($n_{\text{obs}} = 1{,}620$ item×rubric cells). Each marker is a judge's aggregate $r$, coloured by family; the pale bar behind it spans that judge's four per-rubric values. The dotted line marks gpt-5.5 (0.123) and the dashed line the best calibrated 24B judge (0.184). Fine-tune and pref-optim versions (v1/v2/v3) are labelled separately.

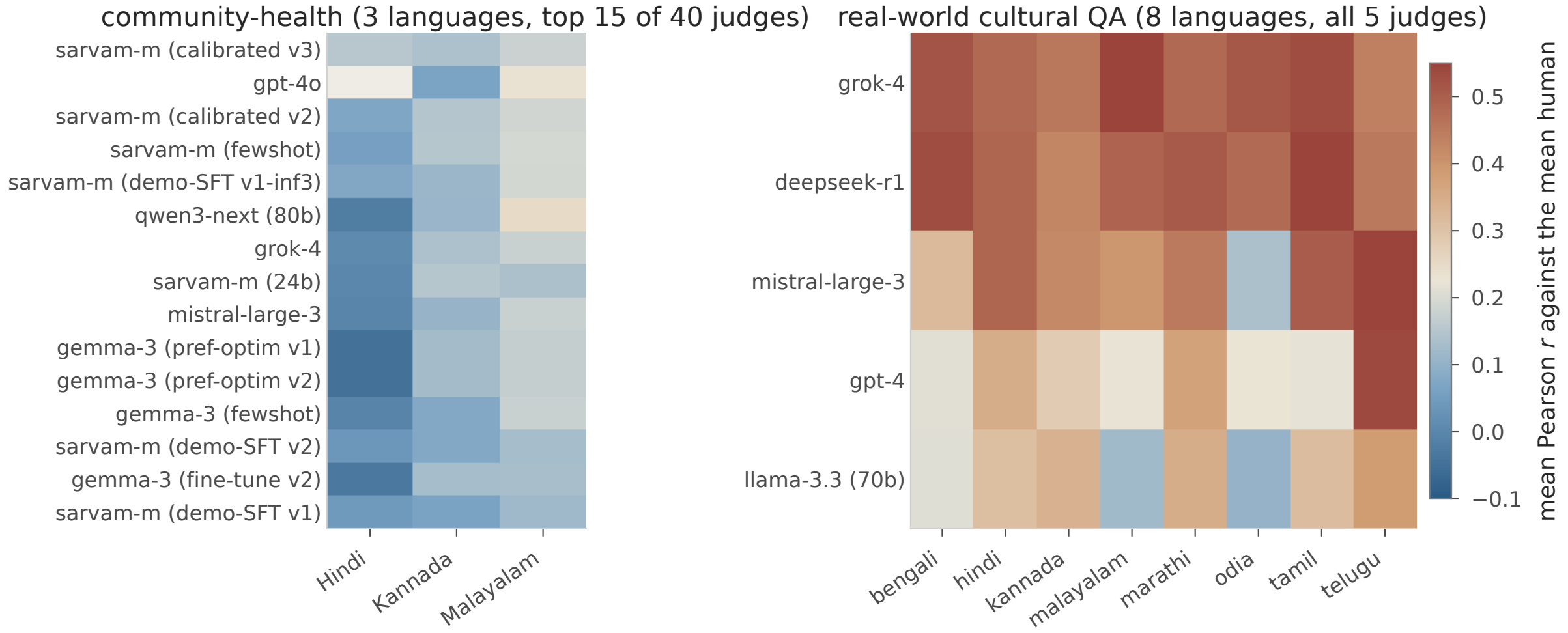

Figure 14: **Per-language agreement, judge by judge.** Mean Pearson $r$ between each judge and the mean-human reference, averaged over that dataset's rubrics, with rows ordered by judge mean and both panels on one colour scale. COMMUNITY-HEALTH (3 languages; the top 15 of its 40 judges, so the rows stay legible) on the left, REAL-WORLD CULTURAL QA (8 languages, all 5 judges) on the right. Agreement varies more across languages within a judge than it does across judges within a language. Per-domain and rubric-kind contrasts are in Fig. 6 and Table 1A.

| Recipe | Applied to | What & key hyper-parameters |
|---|---|---|
| few-shot (FS) | all three families | up to $k=3$ demos per (language × rubric), drawn from the held-out human-label pool, shortest-first with rating-class round-robin; frozen weights; auto-shrink $3\to2\to1$ under the token budget; same decoding as base. |
| fine-tune (FT) | all three families | per-rubric LoRA SFT (Hu et al., 2022; Ouyang et al., 2022) on the 5,683-label corpus; LoRA rank 16–64 ($\alpha = 2r$), target $\{q, k, v, o, \text{gate}, \text{up}, \text{down}\}$, AdamW, cosine schedule, bf16, 2–6 epochs with best-checkpoint selection on eval loss (e.g., sarvam-m v2: $r$=64, $\alpha$=128, best checkpoint at $\approx$1.9 epochs). |
| pref-optim (PO) | all three families | DPO (Rafailov et al., 2023) from the base model on the 2,935 strict-preference triples; v1: $r$=16/$\alpha$=32, $\beta$=0.1, LR $8\times10^{-6}$, 1 epoch; v2: $r$=32/$\alpha$=64, $\beta$=0.05, LR $5\times10^{-7}$, 2 epochs. |
| FT + PO | all three families | stage-wise: merged FT checkpoint → DPO; dominant weight-updating signal in our repair analysis. |
| demo-SFT (AT) | sarvam-m only | demonstration-augmented SFT: $K$ human-rated demos baked into every training prompt (self-demo excluded); AT-v1 ($K$=1), AT-v2 ($K$=3), AT-v1-inf3 (trained with 1 demo, scored with 3-demo prompts). |
| calibrated (CAL) | sarvam-m only | demonstration-anchored calibration: the *frozen* best DPO checkpoint scored with $K$ human-rated demonstrations in context at inference (CAL-v2: $K$=2; CAL-v3: $K$=3); no weight updates and no output-layer rescaling (a pure post-hoc affine rescale is analysed separately in App. M and does not reproduce the gains). |

Table 15: Trained judge variant recipes. All variants share the base-judge JSON schema and decoding settings; demonstration pools are anchored to human-rated items disjoint from the evaluation questions.

| Regime | Rubrics in this study | Measured behaviour |
|---|---|---|
| small-$d_R$ (objective) | DLQ task quality ($d_R\approx$2.1) | judges' dominant shared axis is close to the human one (47.7°); matched per-rubric angle 60.7° against a 56.4° floor, with the best judge inside it; LLM–human agreement reaches 88% of the H–H band (Sec. 5.2). |
| large-$d_R$ (subjective) | CH all four (assigned by rubric type; $d_R$ is measurable only where multiple raters exist); DLQ LA, HAL ($d_R \approx$ 2.3–2.5) | $\sigma$-compression, low effective rank, angles above the inter-human floor yet below the permutation null, and inter-judge clustering, i.e., the four signatures of Sec. 3, quantitatively confirmed with the controls of Sec. 5.4. |

Table 16: Boundary condition: which rubrics fall in which regime and what is measured there. The lens predicts *which* rubrics fail, the *shape* of failure, and *where* it vanishes; it does *not* predict which repair technique closes the gap, a dissociation Sec. 5.7 measures.

**Participation ratio.** $d_R$ (Eq. 1) is computed on the centered items×raters matrix of the three DLQ humans, per rubric×domain cell; no LLM output enters it.

## M Robustness and control tables

### M.1 Per-rubric controls

Table 17 reports, per dataset and rubric, the Pearson/Spearman pair, the raw, rank-transformed and remapped angles, and the permutation null.

### M.2 Human-reference stability

Table 18 varies the human reference construction and shows the objective-below-subjective ordering never flips.

### M.3 Judge-level permutation results on CH

### M.4 Quadratic-weighted $\kappa$

Table 20 repeats the headline agreement comparison under quadratic-weighted $\kappa$ instead of Pearson.

### M.5 Per-language human–human references (DLQ)

Table 21 gives the inter-human reference per language, showing the floor is measurable in every one.

### M.6 Unrecoverable human variance

Cross-validated ridge regression (5-fold, $\alpha$=1, standardised inputs) reconstructing the mean-human score from the full judge cohort: CV $R^2$ = 0.446 on task quality (55.4% unrecoverable), 0.203/0.210 on linguistic acceptability/hallucinations (79.7%/79.0% unrecoverable), and $\leq 0$ on all four CH rubrics ($\approx$ 100% unrecoverable; $n$=405,

| Dataset | Rubric | $r$ | $\rho$ | $\theta_{\text{raw}}$ | $\theta_{\text{rank}}$ | $\theta_{\text{remap}}$ | Null | <null |
|---|---|---|---|---|---|---|---|---|
| DLQ | task quality | 0.519 | 0.503 | 58.5 | 59.6 | 58.7 | 88.7 | 5/5 |
| DLQ | linguistic acceptability | 0.273 | 0.286 | 74.0 | 73.2 | 73.9 | 88.7 | 5/5 |
| DLQ | hallucinations | 0.316 | 0.309 | 71.6 | 72.0 | 71.6 | 88.6 | 5/5 |
| CH | clarity | 0.081 | 0.077 | 85.4 | 85.6 | – | 87.7 | 6/28 |
| CH | helpfulness | 0.103 | 0.093 | 84.1 | 84.7 | 84.4 | 87.7 | 14/39 |
| CH | accuracy | 0.112 | 0.113 | 83.6 | 83.5 | – | 87.7 | 21/40 |
| CH | completeness | 0.121 | 0.118 | 83.0 | 83.2 | – | 87.7 | 21/37 |

Table 17: **Ordinal and permutation controls, per rubric.** $r/\rho$ = mean |Pearson|/|Spearman| over judges vs. the human reference; $\theta_{\text{raw/rank/remap}}$ = mean angle on raw scores, rank-transformed scores, and after monotonically remapping each judge onto the human score marginal; Null = permutation-null mean (500 shuffles); <null = judges below the null at $p < 0.05$ (of judges with $\geq 50$ complete items and non-degenerate variance; "–" = remap undefined for degenerate judge marginals). Spearman tracks Pearson to within 0.02 (max gap 0.017, task quality) and the remap moves angles by $\leq 0.13°$ on DLQ ($\leq 0.31°$ on CH), so compression is not conservative scale use. Tabulated angles are rounded to $0.1°$, while the remap deltas are computed before rounding (task quality: $58.54 \rightarrow 58.67$, i.e. $0.13°$). Cohorts differ from Table 1A, which reports the 14 CH base judges: the <null column here spans up to all 40 CH judges, so its CH angles (83.0–85.4, mean 84.0) and $r$ (mean 0.104) differ slightly from the base-only values there (84.4, 0.093). Human floors quoted here are *pairwise* (61.7/69.1/68.6°); the reference-matched floors used in the body are 56.4/64.4/63.8°. The rubric-level pattern is itself a finding: the CH collapse is strongly rubric-dependent, worst on clarity (6/28 below null) and mildest on accuracy and completeness (21/40, 21/37).

| Reference construction | Objective $\theta$ | Subjective $\theta$ |
|---|---|---|
| 3-human mean | 58.5 | 72.8 |
| single $h_1$ | 65.8 | 77.1 |
| single $h_2$ | 66.5 | 78.2 |
| single $h_3$ | 63.2 | 76.0 |
| leave-out $h_1$ | 60.2 | 74.0 |
| leave-out $h_2$ | 60.2 | 73.7 |
| leave-out $h_3$ | 61.7 | 75.0 |

Table 18: **Human-reference stability (DLQ, per-rubric angles).** Across all seven constructions the objective angle stays $\leq 66.5°$ and the subjective mean stays $\geq 72.8°$: the objective-below-subjective ordering never flips. The stacked 3-rubric analogue varies by only $1.83°$ across constructions.

14 base judges). This measured magnitude is reported in place of any all-or-nothing "null-space" claim.

## M.7 The $d_R$ regression: cells, stability, and entanglement

Across the nine cells (Fig. 15): Spearman $\rho = 0.68$ ($p = 0.042$; one-sided label-permutation $p = 0.0231$ over 20,000 permutations), Pearson $r = 0.91$, $R^2 = 0.83$. Leave-one-cell-out: $\rho \in [0.55, 0.90]$, $R^2 \in [0.78, 0.94]$: no single cell carries the result.

**The relation is between-regime, not graded.** Pearson (0.91) far exceeding Spearman (0.68) is the tell, and the within-group statistics confirm it: across the six subjective cells alone $\rho = -0.03$, and the three objective cells show no usable trend ($n$=3). What the regression detects is the separation between an objective cluster ($d_R \approx 2.1$, $\theta \approx 60°$) and a subjective one ($d_R \approx 2.4$, $\theta \approx 74°$), not a dose–response curve inside either. We therefore report $d_R$ as a regime separator and quote no degrees-per-dimension slope: fitting one over a $d_R$ range of 2.10–2.52, four tenths of a single dimension, invites an extrapolation the data cannot support.

**The measure is capped by the panel size.** With three raters the participation ratio of an items×raters matrix cannot exceed 3, so $d_R$ measures the dimensionality of *rater disagreement as three raters can express it*, not the intrinsic dimensionality of the human quality manifold. A larger panel would be needed for the latter, and would likely shift the absolute values; what a three-rater panel does support is the ordering, and that is all we use.

**Entanglement caveat.** At $n$=9 cells, $d_R$, inter-rater agreement, and label entropy are tightly coupled facets of human disagreement; agreement and entropy each predict the angle about as strongly ($\rho = -0.78, -0.80$), and rank-based partial correlations show the $d_R$ signal does not survive conditioning on rater agreement (partial $\rho = -0.46$) and shrinks under conditioning on entropy (partial $\rho = 0.12$). The supported claim is therefore predictive, not causal, and coarse rather than fine: *a quantity computable from human raters alone, before any LLM is run, tells you which regime a rubric is in.* Alternative readings (label noise, item ambiguity, pretraining familiarity, prompt compatibility)

| Below null | $\theta$ | $p$ | $q_{\mathrm{BH}}$ | Not below null | $\theta$ |
|---|---|---|---|---|---|
| sarvam-m (FT v1) | 85.1 | .0013 | **.025** | sarvam-m (base) | 89.5 |
| gemma-3 (FT v2) | 85.3 | .0018 | **.025** | gemma-3 (base) | 89.3 |
| sarvam-m (FT+PO v2) | 85.3 | .0022 | **.025** | aya-expanse (base) | 89.3 |
| gpt-5.5 | 86.0 | .0025 | **.025** | krutrim-2 | 89.6 |
| llama-3.3 (70b) | 85.1 | .0070 | **.047** | grok-4 | 89.9 |
| gpt-4o | 85.0 | .0071 | **.047** | deepseek-r1 | 89.2 |
| sarvam-m (FT v2) | 86.1 | .0090 | .051 | qwen3-235b | 89.3 |
| gemma-3 (FT+PO v2) | 86.4 | .0155 | .078 | qwen3-next (80b) | 89.6 |
| sarvam-m (CAL v3) | 87.0 | .0193 | .081 | llama-4-scout | 89.7 |
| mistral-large-2 | 86.3 | .0203 | .081 | mistral-large-3 | 88.7 |
| sarvam-m (FT+PO v1) | 86.8 | .0288 | .105 | | |

Table 19: **CH stacked angles vs. a** $20{,}000$**-shuffle permutation null, per judge** (null mean $88.8°$). Exact $p$ with Benjamini–Hochberg (Benjamini and Hochberg, 1995) $q$ across all 40 tests; **bold** $q$ survives $q < .05$. 11 judges are below at $p < .05$ (2 expected by chance), 6 at $q < .05$ and 10 at $q < .10$. The published 300-shuffle version could not attain $q < .05$ for *any* judge, since its smallest possible $p$ $(1/301)$ exceeds the rank-1 BH threshold $.05/40$; that test was resolution-limited, not evidence-limited, which is why we rerun it here. The right column lists the ten *base* judges not below the null; 19 trained variants are also not below and are omitted for space (full table in the release). These are paired tests of each judge against chance, not against each other: see the resolution caveat in Sec. 5.4.

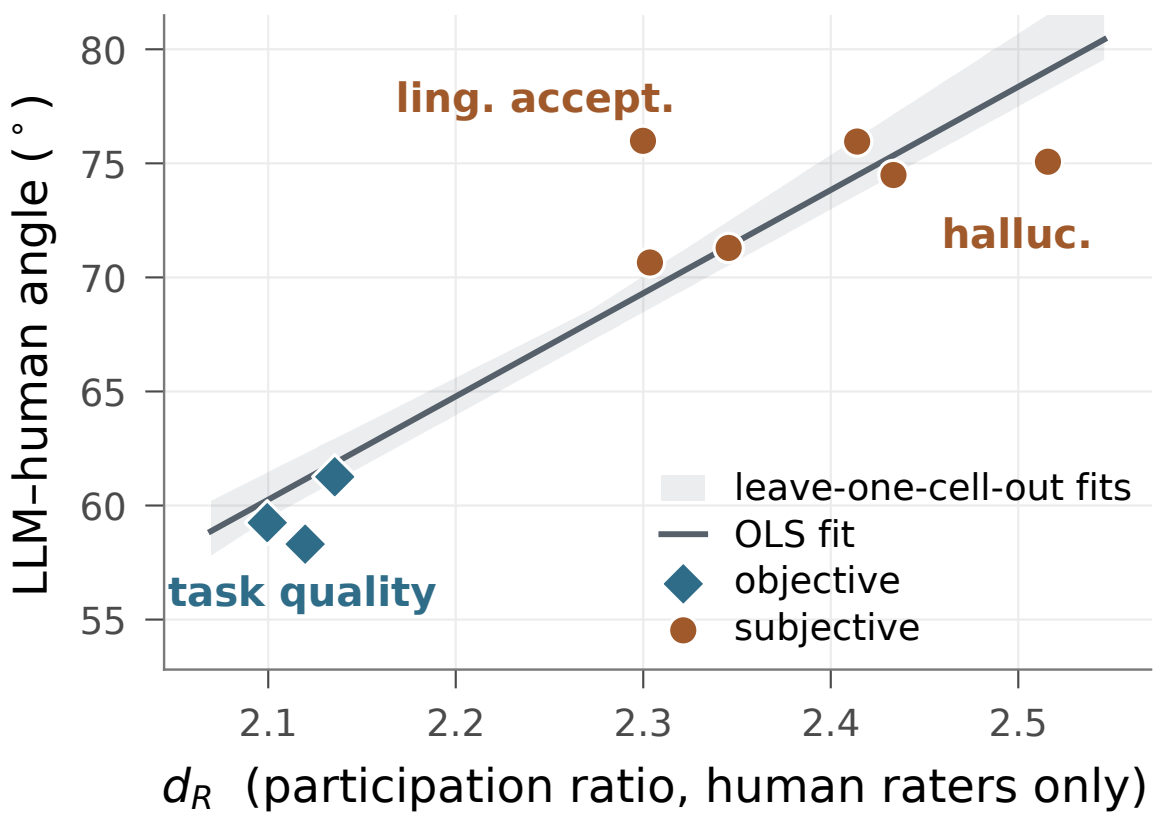

Figure 15: **The $d_R$ boundary predictor** over the nine DLQ rubric×domain cells of Table 22. $d_R$ (Eq. 1) is computed from the three human raters alone, before any LLM output is used; the angle is the mean over the five base judges. The band collects the nine leave-one-cell-out fits, showing that no single cell carries the slope. Objective cells occupy the low-$d_R$, low-angle corner and subjective cells the high-$d_R$, high-angle one.

| Dataset | Rubric | $\mathrm{QWK}_{\mathrm{LH}}$ | $\mathrm{QWK}_{\mathrm{HH}}$ |
|---|---|---|---|
| DLQ | task quality | 0.339 | 0.464 |
| DLQ | linguistic acceptability | 0.185 | 0.355 |
| DLQ | hallucinations | 0.197 | 0.350 |
| CH | clarity | 0.030 | – |
| CH | helpfulness | 0.048 | – |
| CH | accuracy | 0.062 | – |
| CH | completeness | 0.071 | – |

Table 20: **Quadratic-weighted** $\kappa$ (mean over judge–human pairs; human–human where three raters exist). The ordinal-agreement view reproduces the Pearson pattern of Table 1A: near-human on task quality, clearly sub-human on subjective rubrics, near zero on CH.

| Language | TQ $\theta_{\mathrm{HH}}$ / $r_{\mathrm{HH}}$ | LA $\theta_{\mathrm{HH}}$ / $r_{\mathrm{HH}}$ | HAL $\theta_{\mathrm{HH}}$ / $r_{\mathrm{HH}}$ |
|---|---|---|---|
| bengali | 64.1 / 0.44 | 63.2 / 0.45 | 69.3 / 0.35 |
| hindi | 58.7 / 0.52 | 60.0 / 0.50 | 63.4 / 0.45 |
| kannada | 61.4 / 0.47 | 79.3 / 0.19 | 68.4 / 0.37 |
| malayalam | 62.1 / 0.47 | 65.6 / 0.41 | 57.9 / 0.53 |
| marathi | 51.8 / 0.61 | 73.9 / 0.28 | 59.4 / 0.51 |
| odia | 78.1 / 0.20 | 68.8 / 0.36 | 75.7 / 0.25 |
| tamil | 58.6 / 0.52 | 65.2 / 0.42 | 78.3 / 0.20 |
| telugu | 59.7 / 0.50 | 71.4 / 0.32 | 73.3 / 0.29 |

Table 21: **Per-language inter-human references** (mean pairwise angle° / Pearson over the three rater pairs; item counts per language in Table 8). Per-cell human–human references are noisy at small $n$ (e.g., Odia, $n$=60), which is why per-language claims in Sec. 5.5 are read against these cell-level references rather than one global floor.

remain compatible with the data (see Limitations).

| Rubric | Domain | $n$ | $d_R$ | $\theta$ | $r_{\text{HH}}$ | $H$(labels) |
|---|---|---|---|---|---|---|
| task quality | health | 284 | 2.099 | 59.25 | 0.453 | 1.38 |
| task quality | finance | 323 | 2.120 | 58.31 | 0.453 | 1.39 |
| task quality | everyday | 575 | 2.135 | 61.26 | 0.447 | 1.56 |
| ling. accept. | health | 284 | 2.304 | 70.65 | 0.384 | 1.14 |
| ling. accept. | finance | 323 | 2.300 | 75.99 | 0.376 | 1.03 |
| ling. accept. | everyday | 575 | 2.433 | 74.49 | 0.331 | 1.06 |
| hallucinations | health | 284 | 2.414 | 75.96 | 0.331 | 0.90 |
| hallucinations | finance | 323 | 2.516 | 75.07 | 0.299 | 0.88 |
| hallucinations | everyday | 575 | 2.345 | 71.29 | 0.368 | 0.97 |

Table 22: **The nine DLQ rubric×domain cells** behind the $d_R$ regression: participation ratio $d_R$ (from the three human raters alone), mean LLM–human angle $\theta$, mean inter-rater Pearson $r_{\text{HH}}$, and mean human label entropy $H$ (bits). $\theta$ here is measured against the *three-rater mean* reference, so these values run below the reference-matched angles quoted in the body (Table 4). $d_R$ and $\theta$ are given to three and two decimals respectively because the within-subjective-group correlation ($\rho = -0.03$) is not recoverable from coarser rounding: two cells tie at $d_R = 2.30$ and two at $\theta = 76.0$, which alone would flip the sign.

### M.8 Self-preference by answer generator (CH)

CH answers come from three generators (llama-3.1-405b, qwen3, sarvam-m), and two of those families also serve as judges; sarvam-m supplies the calibrated judge behind the training headline. If judges favoured their own family's answers, that alone could produce the result.

Within each (language, question) triple all three generators answer the same question, so we standardise each rater's scores per rubric over the 405 core items, take each judge's deviation from its own triple mean, subtract the human's deviation on the same triple (removing item difficulty and any real quality difference between generators), and then subtract the mean effect of all judges *outside* that family (removing anything about the generator that the whole cohort sees). The residual is the family-specific preference. Against a permutation null that reassigns the same number of judges to each generator (20,000 draws), no family shows an effect: sarvam-m $+0.028$ SD ($p = 0.58$), qwen3 $-0.035$ SD ($p = 0.71$), llama $-0.008$ SD ($p = 0.95$). Self-preference is not a plausible explanation for any result here.

### M.9 Split-half reliability of angles

Random item halves (200 draws): per-rubric DLQ angles agree across halves to a mean $|\Delta\theta|$ of 2.3–2.5°, and the *ordering* of judges by angle is preserved across halves (mean cross-half correlation 0.96/0.97/0.74 on task quality/linguistic acceptability/hallucinations). On CH the mean cross-half $|\Delta\theta|$ is 2.2° and the cross-half *ordering* of the 40 judges is uninformative ($r \approx 0$), because the cohort is uniformly collapsed into an 85–90° band: the between-judge SD is 1.6° while a per-judge bootstrap interval is 6.3° wide. The coarser below-null *binary* does not rescue this at half size either: repeating the full permutation test inside each half (30 draws, 300 shuffles per half) recovers about 6 below-null judges per half but the two halves agree on *which* ones at $\kappa = 0.02$. The full-item test does have power (6 judges survive FDR at $q < .05$ on all 405 items), so this is a budget statement: resolving individual judges in a collapsed regime needs the full item set, not half of it. The study accordingly makes no CH claim that depends on angle ordering; CH judge comparisons use stacked Pearson, whose leaderboard is bootstrap-stable (Sec. 5.7).

### M.10 Ensembling: how far can averaging go?

An ensemble mean must be read against a human reference built the same way. Scoring a $k$-judge mean against the three-rater human mean while scoring a human against a single colleague compares an averaged quantity with an unaveraged one, and on that mismatched pairing the five-judge ensemble appears to sit *below* the human floor on all three rubrics, which would say that five LLMs approximate the human consensus better than one human does. On the matched construction, where the held-out human and the ensemble are both scored against the same two-rater mean, the picture reverses on exactly the rubrics the paper's thesis concerns; Table 3 gives the curve.

Two readings follow. First, ensembling is genuinely useful: it buys 7.6–9.5° between one judge and five. Second, what it buys is variance reduction toward the cohort's dominant axis, not a new direc-tion: the $k$=5 angles land within $\sim 3.5^\circ$ of the cohort PC1-to-human angles ($47.7/64.3/65.5^\circ$), and the extrapolated limits ($66.5^\circ$, $64.6^\circ$) remain outside the subjective floors ($64.4^\circ$, $63.8^\circ$). Where the shared axis misses the human axis, averaging more judges from the same cohort converges to the miss. On CH the ensemble reaches $77.6$–$80.9^\circ$, far outside anything the single merged reference can be compared against.

We flag one asymmetry that runs *against* us and is worth stating: even in the matched construction the ensemble averages five judges while the human side is a single rater. The subjective-rubric conclusion therefore holds despite giving the ensemble the advantage, and the objective-rubric crossing is correspondingly easier to achieve than the table makes it look.

### M.11 Post-hoc rescaling vs. demonstration anchoring

Classic post-hoc calibration rescales scores against labelled anchors (Platt, 1999; Guo et al., 2017); we test whether such rescaling explains the demonstration-anchored gains. The curve is in the body (Fig. 11) and the per-anchor values in Table 23. Fits are per-rubric affine maps estimated on $n$ human-anchored items and evaluated on the held-out remainder, over 500 resamples (seed 20260828); at $n$=5 the gemma-3 fits are degenerate and are omitted. Below roughly 40 anchors the 95% interval spans zero, so rescaling on a small anchor set is not merely weak but unreliable, whereas the demonstration-anchored judge uses the same human-rated items *at inference* and reaches $0.184$.

### M.12 The excluded DLQ comparative split

The DLQ comparative split (21,610 pairs, 10 languages) has exactly one LLM judge (gpt-4, from the source benchmark). Its labels are strongly *anti*-correlated with all three humans ($\kappa = -0.193/-0.197/-0.187$) while the humans agree with each other ($\kappa = 0.542/0.525/0.544$), a pattern consistent with an A/B label-orientation artefact in the inherited judgements rather than genuine anti-alignment. With a single judge and a suspect orientation, the split cannot support any LL-vs-LH comparison, so it is excluded from analysis and counted only in event totals.

## N Reconciling per-rubric and stacked angles

The paper uses four angle constructions, which sit at systematically different magnitudes for the same cohort. Two axes vary independently: *what* is compared (one rubric vector, a whole subspace, or the cohort's dominant axis) and *against which reference* (a single rater, a two-rater mean, or a three-rater mean). Body claims fix both. Table 4 gives the four constructions side by side; the per-construction reading of "what it measures" is as follows. The per-rubric matched construction isolates one rubric's score direction with judge and held-out human on the same target. The three-rater-mean variant is the same comparison, except the judge gets a less noisy target than any individual human does. The per-judge stacked statistic reports the *worst-matched* direction pair between the two 95%-energy subspaces; the cohort stacked variant does the same for the judges pooled into one subspace rather than any single judge. The cohort PCA axis reports the alignment of the judges' *best* shared direction. Human counterparts exist only on DLQ, which has three raters; CH merges its annotators into a single reference, so no floor is measurable there and CH numbers are never compared with one.

The orderings are forced, not accidental: the stacked statistic is the *largest* principal angle, i.e., the worst-aligned pair of directions once each side's 95%-energy basis is taken, so it upper-bounds the picture; the cohort-PCA angle takes the judges' single dominant axis, the best-case direction, so it lower-bounds it; per-rubric angles sit in between because each rubric mixes the judge's aligned dominant component with its misaligned remainder. Concretely, on DLQ task quality the judges' *best shared* direction is well aligned ($47.7^\circ$), each rubric vector moderately so ($60.7^\circ$), while the full subspaces still contain directions nearly orthogonal to the human ones ($84.3$–$88.9^\circ$ stacked), as do the humans' own ($73.9$–$78.1^\circ$): high stacked angles alone do not distinguish judges from humans, which is why every headline contrast is drawn *within* one construction against its own matched human reference or permutation null.

The reference axis matters as much as the construction. On task quality the same five judges score $0.519$ against the three-rater mean but $0.418$ against a single rater; the inter-human value is $0.474$. Reading the first against the third puts judges above the human floor; reading like against like puts them 12% below it. Every comparison in the body uses the matched pairing, and this is the single change that most affects the paper's numbers relative to the submitted version.

| FT+PO judge | $n$=5 | 10 | 20 | 40 | 80 | 160 | uncal. |
|---|---|---|---|---|---|---|---|
| sarvam-m | 0.044 | 0.063 | 0.086 | 0.112 | 0.131 | 0.144 | 0.136 |
| gemma-3 | – | 0.066 | 0.086 | 0.109 | 0.132 | 0.150 | 0.154 |

Table 23: **Post-hoc rescaling vs. anchor count.** Held-out stacked Pearson $r$ for the values plotted in Fig. 11; rescaling never reaches CAL (0.184) and *hurts* below $\sim$40–80 anchors.

# O Reproducibility and artifact index

**Release.** We release, in our public artifact repository: (i) the two consolidated score CSVs (all 170,890 LLM-judge calls, raw model outputs as JSONL, and all human labels used); (ii) the exact judge prompts (App. I); (iii) every metric, figure, and table generator (`framing_analysis.py`, `extended_analysis.py`, `stage_progression_extended.py`, `analysis_plots/build_artifacts.py`); (iv) the control analyses as one runnable script each (`run_rebuttal_perrubric.py`, `run_rebuttal_robustness.py`, `run_rebuttal_theory_tests.py`); and (v) `camera_ready/build_numbers.py`, which regenerates every roster count, event total, and control statistic in this paper and emits them as the LaTeX macros used in the text, plus `camera_ready/build_control_figs.py` for the two control figures, so each printed number and control plot is reproducible by one command. Analysis code is released under the MIT license; judge outputs and annotations under CC BY 4.0, subject to the source benchmarks' terms (Hamna et al., 2026; Watts et al., 2024).

**API judge versions and call dates.** Table 24 pins every hosted judge to a deployment string and a call date, so the closed-model results can be situated even after those endpoints change.

**Reproducibility checklist.** Table 25 maps each claim class to the script and seed that regenerates it.

**Seeds and determinism.** Training seed 42; item-bootstrap seed 20260525 ($B$=1000); permutation seeds 20260710 (500 per-rubric / 300 stacked shuffles); camera-ready control analyses seed 20260828 (20,000 label permutations; 200 split-half draws; 500 anchor-curve resamples). All local decoding is greedy (temperature 0.0), so local-judge scores are deterministic up to serving nondeterminism.

**Event accounting.** LLM-judge calls: 97,200 CH standalone (40 judges $\times$ 2,430 items, 4 rubric scores per call) + 46,170 CH comparative (19 judges $\times$ 2,430 pairs) + 5,910 DLQ standalone (5 $\times$ 1,182, 3 rubric scores per call) + 21,610 DLQ comparative (1 $\times$ 21,610) = 170,890 calls, i.e., 474,310 rubric-level scores and preference labels; human labels total 87,618 (9,720 CH standalone scores, 2,430 CH preferences, 10,638 DLQ standalone scores, 64,830 DLQ preferences).

# P Qualitative study: objective vs. subjective rubrics

This appendix grounds the quantitative geometry results in concrete items drawn from the evaluation set. We separate items into two regimes:

**Objective rubrics.** Items whose target answer is fact-based, verifiable, or grounded in widely shared cultural knowledge. Phrasing may vary across responders, but the underlying information is stable, so evaluation reduces to factual consistency and concept matching. Human and LLM judges converge sharply on these items (Table 26).

**Subjective rubrics.** Items that require contextual interpretation, reasoning, advisory guidance, or sensitivity to the user's situation. There is no single fixed answer; evaluation depends on relevance, completeness, practical applicability, and contextual appropriateness rather than exact factual match. Healthcare items in particular carry emotional, financial, and access-related context that LLM judges tend to overlook even when responses are medically accurate (Table 27).

**Rubric dimensions assessed.** Responses are evaluated along four axes: (i) *task quality / factuality* (accuracy, expected details, direct address of the question); (ii) *linguistic acceptability* (grammar, word choice, spelling); (iii) *clarity* (well-formed, interpretable phrasing); and (iv) *helpfulness and local relevance* (practical usefulness and fit to the user's situational or local context). The objective regime is dominated by axis (i); the subjective regime engages all four, and the failures in Table 27 concentrate on axis (iv).

| Judge | Deployment / version | Call dates | Notes |
|---|---|---|---|
| gpt-5.5 | `gpt-5.5_2026-04-24` | 2026-05-23 | CH standalone; responses API, no temperature parameter |
| deepseek-r1 | `DeepSeek-R1_1` | 2026-05-23 | CH + DLQ; no temperature parameter |
| grok-4 | `grok-4_1` | 2026-05-23 | CH + DLQ; no temperature parameter |
| mistral-large-3 | `Mistral-Large-3_1` | 2026-05-23 | CH + DLQ |
| llama-3.3-70b (DLQ) | `Llama-3.3-70B-Instruct_5` | 2026-05-23 | API-hosted for DLQ; locally served for CH |
| gpt-4o (CH) | inherited from Hamna et al. (2026) | – | version as documented there |
| gpt-4 (DLQ) | inherited from Watts et al. (2024) | – | version as documented there |
| local judges | HF snapshots, vLLM 0.18.0 | 2026-05-22/23 | bf16, temperature 0.0, top-$p$ 1.0, max 512 tokens (parse-retry: 1024–2048); 8×B200 |
| trained variants | LoRA/PEFT checkpoints (released) | 2026-05-23/24 | training seed 42; recipes in App. J |

Table 24: API deployments and call dates for all judges. Scores from the two source benchmarks are used as released by those studies.

| Item | Status |
|---|---|
| Raw judge outputs | released (JSONL per call + two consolidated CSVs) |
| Human labels | released as used; provenance per the source studies |
| Prompts | verbatim in App. I and in the release |
| Analysis / figure code | released; one command regenerates every number (`build_numbers.py`) and every figure (`build_artifacts.py`) |
| Training configs & logs | per-variant YAMLs, trainer logs, and merged-checkpoint recipes released |
| Seeds | training 42; bootstrap 20260525; permutations 20260710; camera-ready controls 20260828 |
| Compute | 8×B200 (vLLM 0.18.0, bf16) for local judges and LoRA training; controls are CPU-only |
| API versions & dates | Table 24 |
| Licenses | code MIT; outputs/annotations CC BY 4.0, subject to source-benchmark terms |

Table 25: Reproducibility checklist for the release.

**Why this matters for the geometry.** The objective items (Table 26, rows Obj-1–Obj-4) sit on a low-dimensional subspace that LLM judges already cover: the correctness axis is shared, and inter-LLM consensus is genuine consensus with humans. The subjective items (Table 27, rows Subj-1–Subj-4) require additional axes (access, vulnerability, affordability, reassurance) along which LLM judges' scores underweight or do not vary at all. These four items are a qualitative analogue of the principal-angle gap (Sec. 5), consistent with judges sharing a medical-completeness direction while humans also weight others; four hand-picked items cannot establish that decomposition, which is why we treat them as grounding rather than evidence.

## Q Use of AI assistants

We used LLM-based assistants for language polishing, for literature search while surveying related work, for arguing through framing and counter-arguments during drafting, and for writing analysis and plotting code. Every claim, number, citation, and line of code so produced was checked and validated manually by the authors, who take full responsibility for the content of this paper.

| ID | Lang | Question (script + English gloss) | Why human and LLM agree |
|---|---|---|---|
| Obj-1 | Ml | ഒരു ബാങ്ക് അക്കൗണ്ട് തുടങ്ങാനായി ആവശ്യമായ രേഖകൾ എന്തൊക്കെയാണ്? (*What documents are required to open a bank account?*) | Procedural recall with a canonical checklist (identity and address proof, photographs, PAN/Aadhaar). All five judges and all three human raters assign identical full-credit task-quality scores. Exact agreement. |
| Obj-2 | Te | తెలుగు సంప్రదాయంలో ఆడపిల్లల తల్లితండ్రులు పిల్లల మొదటి జన్మకు ఎందుకు డబ్బు చెల్లిస్తారు? (*In Telugu tradition, why do the parents of a girl pay for the first childbirth?*) | Cultural practice grounded in widely shared understanding (e.g., *Janmantara Kanuka*). Responses vary in phrasing but capture the core intent; humans explicitly accept the "gist" as sufficient, and LLM judges score accordingly. High agreement. |
| Obj-3 | Hi | डेबिट कार्ड और क्रेडिट कार्ड में क्या अंतर है? (*What is the difference between a debit card and a credit card?*) | Conceptual but well-defined knowledge: debit draws from the user's account; credit is borrowed and repaid with interest. Whenever responses articulate this contrast, both humans and LLMs rate them correct. Consistent agreement. |
| Obj-4 | Kn | ಉತ್ತರ ಕರ್ನಾಟಕದಲ್ಲಿ ಬಳಕೆಯಲ್ಲಿರುವ ಕನ್ನಡ ಹಾಗು ಬೆಂಗಳೂರು ಕನ್ನಡದ ಮಧ್ಯೆ ಇರುವ ವ್ಯತ್ಯಾಸಗಳೇನು? (*What are the differences between the Kannada spoken in North Karnataka and Bengaluru Kannada?*) | Widely shared dialectal knowledge with stable, verifiable contrasts (vocabulary, address forms, Marathi/Urdu vs. English influence). Responses vary in phrasing but the core contrasts are fixed; all five judges and all three raters assign identical scores. |

Table 26: **Objective rubric, qualitative examples.** For fact-based or canonically known items, LLM judges converge with humans on near-identical scores. Referenced from Sec. 1 and Sec. 5 (boundary analysis).

| ID | Lang | Theme | Question (script + English gloss) | What the LLM emphasises vs. what humans need |
|---|---|---|---|---|
| Subj-1 | Kn | Healthcare access, vulnerability, and power dynamics | ನಮ್ಮನೆ ಹತ್ರ ಇರೋ ಪ್ರಾಥಮಿಕ ಆರೋಗ್ಯ ಕೇಂದ್ರದಲ್ಲಿ ತೋರ್ಸೋಕೆ ಹೋದ್ರೆ ಎಲ್ಲಾದಕ್ಕೂ ದುಡ್ಡು ಕೇಳ್ತಾರೆ ನಾನ್ ಇವ್ರ ಮೇಲೆ ದೂರು ಕೊಡ್ಬೇಕು ಅಂದ್ರೆ ಏನ್ ಮಾಡ್ಬೇಕು? (*If the nearby primary health centre asks for money to be seen, how do I file a complaint against them?*) | **LLM:** detailed procedural guidance on filing complaints. **Human-aligned:** acknowledges fear of retaliation, dependence on the same centre, and recommends supportive pathways (community health workers, local officials, collective reporting). The question is about institutional trust and social hierarchy, not paperwork. |
| Subj-2 | Ml | Medical recommendations vs. real-world accessibility | എന്റെ ഒമ്പത് വയസ്സുള്ള മോൾ വീണ് നെറ്റി മുറിഞ്ഞു, ഒരുപാട് വലിയ പാട് വരാതെ സ്റ്റിച് ചെയ്യാൻ ഏത് വിഭാഗത്തിലാണ് കുട്ടിയെ കാണിക്കേണ്ടത്? (*My nine-year-old daughter cut her forehead in a fall; which department should I take her to so it can be stitched without a large scar?*) | **LLM:** narrowly recommends a plastic surgeon. **Human-aligned:** balances clinical optimality with accessibility and affordability; many cuts are handled effectively by emergency physicians, paediatric surgeons, or general surgeons if treated promptly. The user's real concern is healing and minimal scarring, not access to specialist cosmetic care. |
| Subj-3 | Hi | Overmedicalisation of everyday living | घर में बिल्ली और कुत्ते जैसे पालतू जानवर हैं। क्या उनके संपर्क से बच्चों और बड़ों को एलर्जी हो सकती है? रोज़मर्रा की ज़िंदगी में क्या एहतियात रखनी चाहिए? (*We keep pets like cats and dogs at home. Can contact cause allergies in children and adults? What everyday precautions should we take?*) | **LLM:** foregrounds HEPA filters, specialised vacuum cleaners, furniture replacement. **Human-aligned:** starts from low-cost, household-level precautions (cleanliness, handwashing, ventilation, symptom monitoring) and escalates only if needed. The user wants to keep living with pets, not to medicalise their home. |
| Subj-4 | Ml | Emotional reassurance and risk framing in maternal health | 6 വർഷം മുമ്പ് ഞാൻ എന്റെ മകളെ പ്രസവിച്ചത് സിസേറിയൻ ആയിരുന്നു. ഇനി ഒരു കുട്ടിക്ക് ശ്രമിക്കേണ്ട സമയം ആയെന്ന് എല്ലാവരും പറഞ്ഞു തുടങ്ങി. ഞങ്ങളും അതേ തീരുമാനത്തിലാണ്. എന്നാലും എനിക്ക് ഇനി സുഖ പ്രസവം ആവാൻ ആണ് ആഗ്രഹിക്കുന്നത്. അതിനു എന്തെങ്കിലും ബുദ്ധിമുട്ടുകൾ ഉണ്ടാവുമോ? എന്തൊക്കെയാണ് അതിനു വേണ്ടി മുൻകരുതേണ്ടത്? (*My first delivery 6 years ago was a Caesarean. Family is asking us to try for a second child and we are deciding the same; I want a normal delivery this time. Are there difficulties? What precautions should I take?*) | **LLM:** foregrounds uterine rupture, foetal distress, and worst-case complications under VBAC. **Human-aligned:** first validates the desire for a normal delivery, notes that many VBACs succeed under proper care, then discusses risks proportionately. The user is asking for reassurance and confidence under family pressure, not a complications checklist. |

Table 27: **Subjective rubric, healthcare qualitative examples.** On items that carry institutional, financial, or emotional context, LLM judges score on medical-completeness while humans weight practical accessibility, vulnerability, and reassurance. These items illustrate, rather than measure, the principal-angle gap reported in Sec. 5; a systematic per-axis test would need item-level coding of the community-relevance axes, which we do not attempt here. Referenced from Sec. 1 and Sec. 5.2.